\documentclass{egpubl}
\usepackage{eurovis2026}

\SpecialIssuePaper         %

\CGFccby

\usepackage[T1]{fontenc}
\usepackage{dfadobe}  

\usepackage{cite}  %

\BibtexOrBiblatex
\electronicVersion
\PrintedOrElectronic
\ifpdf \usepackage[pdftex]{graphicx} \pdfcompresslevel=9
\else \usepackage[dvips]{graphicx} \fi

\usepackage{egweblnk}

\title[Manifold-Preserving Superpixel Hierarchies and Embeddings]%
{Manifold-Preserving Superpixel Hierarchies and Embeddings \\for the Exploration of High-Dimensional Images}

\makeatletter 
\let\mainPaperTitle\@title 
\makeatother

\author[A. Vieth et al.]
{\parbox{\textwidth}
	{\centering 
        A. Vieth$^{1}$\orcid{0000-0002-5809-4316}, %
        B. Lelieveldt$^{1}$\orcid{0000-0001-8269-7603}, %
        E. Eisemann$^{2}$\orcid{0000-0003-4153-065X}, %
        A. Vilanova$^{3}$\orcid{0000-0002-1034-737X} and %
        T. Höllt$^{2}$\orcid{0000-0001-8125-1650}
    }
    \\
	{\parbox{\textwidth}
		{\centering %
			         $^1$Leiden University Medical Center, Leiden, The Netherlands
			         $^2$Delft University of Technology, Delft, The Netherlands \\
			         $^3$Eindhoven University of Technology, Eindhoven, The Netherlands \\
        }
    }
}

\usepackage{lipsum}  

\usepackage{amssymb, amsfonts}  %
\usepackage{amsmath}            %
\usepackage{enumitem}           %
\DeclareMathAlphabet{\dutchcal}{U}{dutchcal}{m}{n}
\usepackage{siunitx}            %
\usepackage{subcaption}         %
\usepackage{booktabs}           %
\usepackage[referable]{threeparttablex} %
\usepackage{multirow}           %
\usepackage[table]{xcolor}      %
\usepackage{dblfloatfix}        %
\usepackage{cuted}              %

\makeatletter
\newcommand\Autoref[1]{\@first@ref#1,@}
\def\@throw@dot#1.#2@{#1}%
\def\@set@refname#1{%
    \edef\@tmp{\getrefbykeydefault{#1}{anchor}{}}%
    \xdef\@tmp{\expandafter\@throw@dot\@tmp.@}%
    \ltx@IfUndefined{\@tmp autorefnameplural}%
         {\def\@refname{\@nameuse{\@tmp autorefname}s}}%
         {\def\@refname{\@nameuse{\@tmp autorefnameplural}}}%
}
\def\@first@ref#1,#2{%
  \ifx#2@\autoref{#1}\let\@nextref\@gobble%
  \else%
    \@set@refname{#1}%
    \@refname~\ref{#1}%
    \let\@nextref\@next@ref%
  \fi%
  \@nextref#2%
}
\def\@next@ref#1,#2{%
   \ifx#2@ and~\ref{#1}\let\@nextref\@gobble%
   \else, \ref{#1}%
   \fi%
   \@nextref#2%
}
\makeatother

\newcommand\checkMaxPages[1]{
\ifnum\thepage>#1
    \noindent
    \colorbox{orange}{
    \parbox{\columnwidth}{
    MAX #1 PAGES ALLOWED FOR CONTENT \\
    (including all images but excluding references.) \\
    This is page \thepage. That's too much. \\
    See \href{https://www.eg.org/wp/eurographics-publications/guidelines/}{CGF AUTHORS’ GUIDELINES}.
    }}
\fi
}

\usepackage{xparse}
\usepackage[user,hyperref]{zref}
\makeatletter
\providecommand{\@currentshorttitle}{}
\zref@newprop{shorttitle}{\@currentshorttitle}
\zref@addprop{main}{shorttitle}

\NewDocumentCommand{\labelshort}{om}{%
  \begingroup
  \IfValueT{#1}{%
    \renewcommand{\@currentshorttitle}{#1}%
    \zlabel{#2}%
  }%
  \endgroup
  \label{#2}%
}

\NewDocumentCommand{\nameshortref}{O{}m}{%
  \zref@ifrefundefined{#2}{%
  }{%
    \hyperlink{\zref@extract{#2}{anchor}}{#1\zref@extract{#2}{shorttitle}}%
  }%
}
\makeatother

\usepackage{cleveref}	%

\newcommand{\imgGraph}{\ensuremath{\mathcal{I}}}
\newcommand{\imgHighDim}{\ensuremath{f}}
 
\newcommand{\imgX}{\ensuremath{X}}

\newcommand{\imgY}{\ensuremath{Y}}

\newcommand{\imgSizeXY}{\ensuremath{|X \times Y|}}
\newcommand{\imgSizeTot}{\ensuremath{n}}

\newcommand{\posx}{\ensuremath{x}}
\newcommand{\posy}{\ensuremath{y}}

\newcommand{\graph}{\ensuremath{\mathcal{G}}}
\newcommand{\edges}{\ensuremath{\mathcal{E}}}
\newcommand{\edgesG}{\ensuremath{\mathcal{E}^{\prime}}}
\newcommand{\vertices}{\ensuremath{\mathcal{V}}}

\newcommand{\edge}[1][\mathbf{ij}]{\ensuremath{e}_{#1}}
\newcommand{\vertex}[1][\mathbf{i}]{\ensuremath{v}_{#1}}

\newcommand{\superpixel}{\ensuremath{\mathcal{S}}} %
\newcommand{\component}{\ensuremath{\dutchcal{s}}} %
\newcommand{\numSuperpixel}{\ensuremath{s}}

\newcommand{\superpixelGroundTruth}{\ensuremath{\mathcal{T}}}
\newcommand{\componentGroundTruth}{\ensuremath{\dutchcal{t}}}

\newcommand{\imgPos}[1][i]{\ensuremath{\mathbf{#1}}}
\newcommand{\id}[1][i]{\ensuremath{#1}}
\newcommand{\ID}[1][i]{\ensuremath{\mathbf{#1}}}

\newcommand{\numAttrDimensions}{\ensuremath{C}}

\newcommand{\dataPointHighDim}{\ensuremath{\mathbf{a}}}
\newcommand{\dataPointScalar}{\ensuremath{a}} %

\newcommand{\randomWalkNumber}{\ensuremath{\omega}}
\newcommand{\randomWalkSteps}{\ensuremath{\lambda}}

\newcommand{\transitions}{\ensuremath{T}}

\newcommand{\mean}{\ensuremath{\mu}}

\definecolor{White}{rgb}{1.0, 1.0, 1.0}
\definecolor{WhiteSmoke}{rgb}{0.96, 0.96, 0.96}

\usepackage{calc}
\newcommand{\zerospace}{\hspace{\widthof{0}}}

\usepackage{tikz}
\usepackage{xcolor}
\usetikzlibrary{shapes.geometric}

\definecolor{GreenDot}{HTML}{22b473}
\definecolor{BlueDot}{HTML}{29a9e1}

\NewDocumentCommand{\inlineCircle}{ O{black} O{} O{6pt} O{12pt}}
{
\kern-2pt
	\resizebox{#3}{!}{
        \begin{tikzpicture}
            \draw[draw=#1, dash pattern=#2, line width=#4](-1,0) circle (1);
        \end{tikzpicture}
	}
\kern-2pt
}

\NewDocumentCommand{\inlineDisk}{ O{white} O{black} O{8pt} O{2pt}}
{
\kern-2pt
	\resizebox{#3}{!}{
        \begin{tikzpicture}
            \filldraw[fill=#1, draw=#2, line width=#4](-1,0) circle (1);
        \end{tikzpicture}
	}
\kern-2pt
}

\newcommand\inlineRect[2][8pt] 
{
\kern-4pt
	\resizebox{#1}{!}{
        \begin{tikzpicture}
            \fill[fill=#2](0.1,0.1) rectangle (0.4,0.4);
        \end{tikzpicture}
	}
\kern-4pt
}

\NewDocumentCommand{\inlineRectangle}{ O{black} O{white} O{8pt} O{2pt}}
{
\kern-4pt
	\resizebox{#3}{!}{
        \begin{tikzpicture}
            \filldraw[draw=#1, fill=#2, line width=#4](0.1,0.1) rectangle (0.4,0.4);
        \end{tikzpicture}
	}
\kern-4pt
}

\newcommand{\smallsim}{\smallsym{\mathrel}{\sim}}
\makeatletter
\newcommand{\smallsym}[2]{#1{\mathpalette\make@small@sym{#2}}}
\newcommand{\make@small@sym}[2]{%
  \vcenter{\hbox{$\m@th\downgrade@style#1#2$}}%
}
\newcommand{\downgrade@style}[1]{%
  \ifx#1\displaystyle\scriptstyle\else
    \ifx#1\textstyle\scriptstyle\else
      \scriptscriptstyle
  \fi\fi
}
\makeatother

\ccsdesc[500]{Human-centered computing~Visual analytics}
\ccsdesc[300]{Mathematics of computing~Dimensionality reduction}
\ccsdesc[100]{Computing methodologies~Image segmentation}

\begin{document}

\teaser{
\centering
\vspace{-1mm}
\includegraphics[width=\linewidth, alt={A teaser.}]{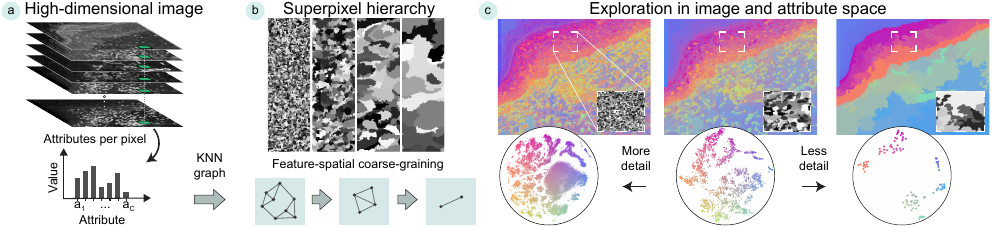}
{\phantomsubcaption\label{fig:teaser:a}}%
{\phantomsubcaption\label{fig:teaser:b}}%
{\phantomsubcaption\label{fig:teaser:c}}%
\vspace{-3mm}
\caption{%
    \textbf{Exploration of high-dimensional images (a) using our superpixel-based hierarchical embeddings.} %
    The hierarchy (b) is based on a $k$-nearest neighbor graph computed on pixel attributes. %
    We coarse-grain this graph, restricted by (super)pixel adjacency, using a similarity derived from random-walks on the neighborhood graph. %
    The graph vertices on each level correspond to superpixels. %
    The same similarities between vertices function as neighborhood indicator to compute embeddings on each hierarchy level (c).
}
\vspace{4mm}
\label{fig:teaser}

}

\maketitle

\begin{abstract}
	High-dimensional images, or images with a high-dimensional attribute vector per pixel, are commonly explored with coordinated views of a low-dimensional embedding of the attribute space and a conventional image representation.
Nowadays, such images can easily contain several million pixels.
For such large datasets, hierarchical embedding techniques are better suited to represent the high-dimensional attribute space than flat dimensionality reduction methods.
However, available hierarchical dimensionality reduction methods construct the hierarchy purely based on the attribute information and ignore the spatial layout of pixels in the images.
This impedes the exploration of regions of interest in the image space, since there is no congruence between a region of interest in image space and the associated attribute abstractions in the hierarchy. 
In this paper, we present a superpixel hierarchy for high-dimensional images that takes the high-dimensional attribute manifold into account during construction.
Through this, our method enables consistent exploration of high-dimensional images in both image and attribute space.
We show the effectiveness of this new image-guided hierarchy in the context of embedding exploration by comparing it with classical hierarchical embedding-based image exploration in two use cases.

	\printccsdesc   
\end{abstract}  

\section{Introduction}  \label{sec:intro}

\begin{figure*}[t]
    \centering
    \includegraphics[width=1\linewidth]{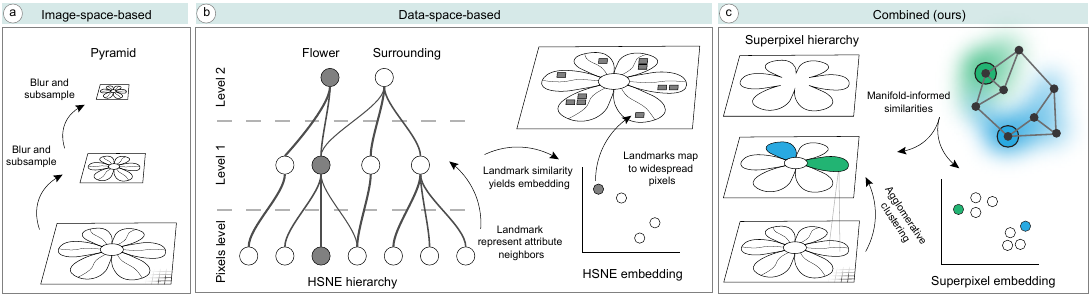}
    {\phantomsubcaption\label{fig:hierarchies:a}}%
    {\phantomsubcaption\label{fig:hierarchies:b}}%
    {\phantomsubcaption\label{fig:hierarchies:c}}%
    \vspace{-4mm}
    \caption{\textbf{Image Hierarchies:} %
    Classical image-space-based hierarchies, like image pyramids, progressively blur and subsample the image (a). 
    Attribute-space-based hierarchies, as used in hierarchical DR methods (b), ignore the image space entirely and only aim to preserve manifold structure of the attribute space.
    A pixel, as highlighted in the hierarchy, might actually be represented by multiple landmarks in abstraction levels.
    And, in turn, a landmark can represent scattered pixels.
    Combined, this complicates an image-based exploration of the high-dimensional space.
    Our superpixel hierarchy (c) combines both image layout and attribute space manifold structure.
    \inlineDisk[GreenDot][GreenDot][5pt] and \inlineDisk[BlueDot][BlueDot][5pt] color two superpixels on the middle abstraction level, the respective coarse-grained vertices in the attribute graph and corresponding embedding points. 
    }%
    \label{fig:hierarchies}%
    \vspace{-4mm}
\end{figure*}

Many fields %
produce images in which each pixel is associated with a high-dimensional attribute vector representing e.g., spectral information (hyperspectral imaging in geoscience), abundance of many chemical elements (macro XRF in cultural heritage analysis), or expression of many proteins or genes (mass cytometry in system biology).
Exploration and analysis of these high-dimensional images is a multi\-faceted challenge as interesting patterns might be observed in the image domain, in the high-dimensional attribute space, or only by combining both.
Embedding the attribute data into lower-dimensional representations with techniques like \mbox{t-SNE}~\cite{VanderMaaten2008, VanderMaaten2014tSNE} and UMAP~\cite{McInnes2018UMAP} has proven useful for visualization and insight generation, since direct visualization is not possible as it would be for traditional color images. %
However, applying these dimensionality reduction (DR) techniques to image data introduces various challenges.
For one, common image sizes of several million pixels push most DR methods to their limits regarding both computational speed and interpretability of the resulting embedding~\cite{Kobak2019}. 
Moreover, standard DR methods do not take the spatial arrangement of image data into account and separate the exploration of image and attribute spaces. 
Recent approaches to overcome this limitation include spatial information by first segmenting the image and subsequently using segment-morphology as part of the embedded features as well as using the segments as glyphs in the embedding representation~\cite{Warchol2025SEAL}.
In this paper, we present a single-step workflow using superpixel hierarchies which directly couples image and attribute space.

Hierarchical DR techniques~\cite{Paulovich2008, Pezzotti16HSNE, Marcilio-Jr2021HUMAP, kuchroo2022} tackle the scalability issues of single-level embedding methods.
They reduce the number of embedded points by creating data hierarchies with increased rates of abstraction in upper hierarchy levels.
Now, instead of embedding the original data, hierarchical DR techniques embed (subsets of) the abstracted or aggregated points, often called landmarks, on a given hierarchy level. 
With each added abstraction level the DR techniques embed fewer landmarks, thus reducing scaling issues.
However, to the best of our knowledge, no hierarchical embedding method exists, that takes the spatial layout of pixels in high-dimensional images into account when constructing the hierarchy.
Thus, landmarks in these methods often represent pixels scattered across the image and, vice versa cohesive regions of pixels with similar attribute vectors are represented by multiple landmarks (\autoref{fig:hierarchies:b}).
As a result, embeddings steered from either the image or embedding space contain more landmarks than needed, leading to increased computational cost and reduced clarity~\cite[Section 6]{Vieth23}.

Here, we propose an extension to hierarchical embeddings which uses a hierarchy that combines aggregation in attribute space and image space through superpixels (\autoref{fig:hierarchies:c}), i.e., arbitrarily shaped, continuous pixel groups which segment the image.
In this context, a superpixel hierarchy is a sequence of superpixel segmentations with increasingly large and consequently fewer superpixels, as shown in~\hyperref[fig:teaser]{Figure 1b}. %
When dealing with classic color images, superpixel segmentation is typically based on a similarity measure between perceptual features like distances in perceptually uniform color spaces.
We adapt this concept to general high-dimensional attribute spaces and combine it with non-linear DR. 
However, these standard measures are unsuitable for general high-dimensional attribute spaces as they fail to capture the underlying non-linear data structure.
We designed a superpixel hierarchy using a similarity measure explicitly tailored to preserve the high-dimensional manifold.
We use this similarity measure for merging superpixels to build the hierarchy and also for estimation of pairwise similarities needed for embedding the data on any hierarchy level.
Specifically, we define this measure in terms of the overlap of visit count distributions derived from random walks on a $k$-nearest neighbor graph of the high-dimensional attributes.
Thereby, and in contrast to other hierarchical DR techniques, our superpixel hierarchy and embedding is informed not only by the high-dimensional attributes but also the spatial component of image data.

We validate our method by two use cases showing the exploration of high-dimensional imaging data using our hierarchical superpixel embeddings, and a quantitative evaluation of the hierarchy.
For the former, we use a hyperspectral satellite imaging dataset and highly-multiplexed tissue imaging (CyCIF) dataset, respectively.
We show that our hierarchical embeddings can represent spatial regions more concisely using fewer landmarks compared to HSNE~\cite{Pezzotti16HSNE}.
Further, we show that our superpixel hierarchy creates landmarks that are semantically meaningful in the spatial context.
Finally, our quantitative hierarchy evaluation, using existing ground truth data, shows that our superpixel hierarchy is of competitive quality, while also allowing for \emph{overview first, detail on demand} exploration of high-dimensional image data.

The main contribution of this paper is connecting the image and attribute space of large high-dimensional images for exploration with hierarchical dimensionality reduction. 
To accomplish this, we \textbf{(1)} designed and implemented a superpixel hierarchy specifically for high-dimensional images, based on \textbf{(2)} a high-dimensional manifold-aware similarity measure between superpixels, and \textbf{(3)} hierarchical neighborhood embedding using the same manifold-aware similarity measure.

\section{Related Work}  \label{sec:relatedWork}

There exist a wide range of \textbf{superpixel methods} and their full discussion is out of scope of this paper; instead, we refer to more extensive reviews~\cite{wangSuperpixelSegmentationBenchmark2017, stutzSuperpixelsEvaluationStateoftheart2018, barcelosComprehensiveReviewNew2024}.
Typically, superpixel methods work with three-dimensional color images, though here we will focus on methods most pertinent to our work, which are those specific to high-dimensional images.

Hyperspectral imaging (HSI) is the main high-dimensional image domain which uses superpixel methods, usually aimed at downstream tasks like pixel classification, endmember detection and hyperspectral unmixing.
Using Felzenszwalb and Huttenlocher's~\cite{Felzenszwalb2004} image segmentation method Thompson et al.~\cite{Thompson2010SuperpixelEndmember} describe a superpixel segmentation methods for hyperspectral images with the goal of endmember detection.
They perform agglomerative clustering to define superpixels and merge clusters based on the Euclidean or cosine distance of the spectra associated with two pixels, comparing minimal pixel-distances within and between superpixels. 
While this approach does create a superpixel hierarchy, it does not preserve the manifold structure of the underlying spectral data.
Similarly adapting an originally color-focused method, Xu et al.~\cite{xuRegionalClusteringbasedSpatial2018} follow the k-means based SLIC~\cite{achantaSLIC2012} and substitute the color distance therein with a domain-specific distance based on cross-entropy and cosine distance. 
Barbato et al.~\cite{barbato22} also adapt SLIC, augmenting it with both a hyperspectral distance and an additional spectral clustering preprocessing step. 
Another noteworthy superpixel method is entropy rate superpixel~\cite{Liu2011Entropy}.
They employ an entropy-based objective function for superpixel segmentation that combines the entropy rate of a random walk on a graph with a balancing term to promote compact, homogeneous, and similarly sized clusters. 
The method constructs a graph in image space by iteratively adding edges between pixels that maximize random walk entropy within the new superpixel. %
Several authors extend this method to HSI: 
whereas Tang et al.~\cite{tangDifferentVersionsEntropy2019} advocate using spectral distances to set up the random walk transition probabilities, others use the first or first three principal components to create a false-color image and apply color distances~\cite{Fan2023}.
The entropy rate superpixel method also creates a superpixel hierarchy, but does not explicitly define distances or similarities between superpixels, which we aim to do in order to embed each hierarchy level.
Grady~\cite{gradyRandomWalksImage2006} describes another method that uses random walks in image space to define merge criteria for superpixels. 
Given specific seed pixels, a pixel is assigned to a superpixel based on which seed pixel is visited most often by random walks started on the pixel itself.
Follow-up papers introduce additional shape constraints and allow self-loops in their walks, as well as data-based seed selection techniques~\cite{shenLazyRandomWalks2014, kangDynamicRandomWalk2020}. %
However, these methods require a fixed number of seed points, and thereby superpixels, and do not build a superpixel hierarchy, in which higher-level superpixels result from merging lower-level superpixels. 
Further, we perform random walks in the attribute space, and not image space, to better preserve the underlying data structure.

Agglomerative methods all implicitly create superpixel hierarchies but usually chose a single hierarchy level as the resulting image segmentation.
In contrast, Wei et al.~\cite{Wei2018SuperpixelHierarchy} explicitly compute all levels of a superpixel hierarchy based on Bor\r{u}vka's algorithm for finding a minimum spanning trees applied to images. %
Yan et al.~\cite{Yan2022HierarchicalSuperpixel} extend this method for asymmetrical distance measures between pixels.
We base our superpixel hierarchy on the adaption of the Bor\r{u}vka's algorithm presented by Wei et al.~\cite{Wei2018SuperpixelHierarchy}.
However, they ignore the manifold structure in the high-dimensional attribute space. 
To inform the superpixel distance and superpixel merging with the data manifold structure we employ random walks in the pixel attribute space, unlike other methods mentioned above which perform random walks in image space.
We additionally link the superpixel generation with exploration of the high-dimensional attribute space via embeddings.

Extensive reviews on dimensionality reduction and multidimensional projection techniques can be found~\cite{Nonato2019, Espadoto2021}.
Most relevant for us are \textbf{hierarchical dimensionality reduction methods}. 
For example, Glimmer~\cite{ingramGlimmerMultilevelMDS2009} (i.e., a hierarchical version of classical multidimensional scaling) and HiPP~\cite{Paulovich2008} perform recursively subsampling and hierarchical clustering respectively to create data hierarchies.
HSNE~\cite{Pezzotti16HSNE} (i.e., a hierarchical version of t-SNE~\cite{VanderMaaten2008}) and HUMAP~\cite{Marcilio-Jr2021HUMAP} (i.e., a hierarchical version of UMAP~\cite{McInnes2018UMAP}) both select landmarks from lower hierarchy levels as points in higher levels and use random walks to define similarities between them.
Multiscale PHATE~\cite{kuchroo2022} (i.e., a hierarchical version of PHATE~\cite{Moon2019}) simulates a diffusion process using random walks to "smoothen" the data and coarse-grain the k-nearest-neighbor data graph repeatedly.
In general, random walks on nearest-neighbor graphs of the attribute data are a popular means to capture manifold-preserving similarities in high-dimensional data.
While it is possible to apply these methods to high-dimensional images, none of them takes the spatial information of images into account when creating hierarchies or defining similarities between sample points.
However, our goal is to merge superpixels in image space based on similarities in pixel attribute space.

\textit{Isomap}~\cite{Tenenbaum2000} introduced the idea of using \textbf{geodesic distances} into DR methods that build on multidimensional scaling to better preserve the manifold structure of the data in the resulting embedding.
The geodesic distances are approximated as the shortest path on a nearest-neighbor graph of the high-dimensional input data.
However, Lee and Verleysen~\cite{leeNonlinearDimensionalityReduction2005} as well as Lafon and Lee~\cite{Lafon2006} discuss that using shortest-path-based similarities can be susceptible shortcut-connections that jeopardize the representation of the underlying data manifold, whereas similarities produced by random walks on a nearest-neighbor graph~\cite{Gobel1974, Crane2020} seem to be more robust, since they consider multiple paths.
Diffusion Maps~\cite{Lafon2006} describes a unified framework of random walk-based dimensionality reduction methods and shows their equivalence to an eigenvector problem. %
Embedding nodes (vertices) from a graph structure is a problem related to dimensionality reduction methods, though such node embedding methods differ from DR used for visualization and exploration in that they are typically designed for the downstream tasks of multi-label classification and link prediction~\cite{khosla2021}.
Both shortest-path, e.g., in~\textit{node2vec}~\cite{groverNode2vec2016}, and random walk-based, e.g., with~\textit{tsNET}~\cite{kruigerGraphLayoutsTSNE2017}, similarity approaches are prominent in these methods as well.

\begin{figure}[b]
    \centering
    \vspace{-1mm}
    \includegraphics[width=0.95\linewidth]{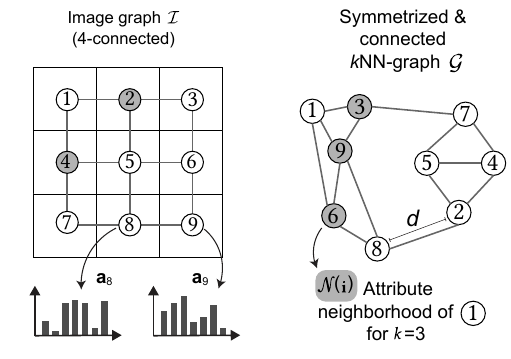}
    \vspace{-1mm}
    \caption{\textbf{Graph structures}: %
    4-connected image graph~\imgGraph\ (left) and attribute-based graph~\graph\ (right).
    Different neighborhoods of the same node \emph{1} are highlighted in grey.
    }%
    \label{fig:notation:graphs}%
    \vspace{-2mm}
\end{figure}

\begin{figure*}[t]
    \centering
    \includegraphics[width=1\linewidth]{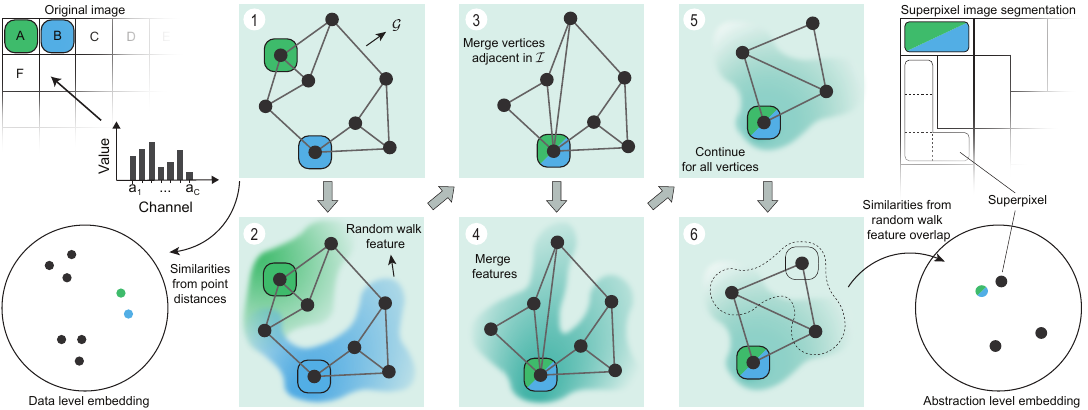}
    \vspace{-3mm}
    \caption{\textbf{Method overview:} %
   	Each image pixel is associated  with a high-dimensional attribute vector (top left). 
   	(1)~We compute a neighborhood graph~\graph, whose vertices correspond to pixels and edges are based on attribute similarities. 
   	The data level embedding is computed from this graph, like in, e.g., t-SNE.
   	(2)~We compute a feature vector per vertex, describing the local graph structure, using random walks. 
   	(3)~For the next abstraction level, vertices in the attribute graph~\graph\ are merged with the most similar neighbor in~\imgGraph. %
   	(4)~The new vertex retains all outgoing connections of the merged vertices.
   	All vertex features are added and re-normalized.
   	(5)~This is repeated for each vertex.
   	(6)~Similarities between merged vertices are used for creating embeddings where
   	each point corresponds to one superpixel (right).
    }%
    \label{fig:Method}%
    \vspace{-3mm}
\end{figure*}

\section{Notation}  \label{sec:method:problem}

We define an image as the set of pixels at coordinates~$\imgX, \imgY \subset \mathbb{N}$, which span the image domain along its two dimensions in a rectilinear grid.
${\imgSizeTot = \imgSizeXY}$ denotes the total number of pixels, and we index a pixel with~%
${\imgPos = (\posx, \posy)_{\id}}$ 
where~%
${1 \leq \id \leq \imgSizeTot}$. 
We can also interpret such an image as a graph, typically four- or eight-connected:~%
\mbox{
$\imgGraph = (\vertices, \edges)$ 
}
where each vertex~%
$\vertex \in \vertices$ 
is an image pixel with edges~%
$\edge \in \edges$ 
between neighboring pixels (\autoref{fig:notation:graphs}, left).
In a four-connected graph, neighboring means the four directly adjacent pixels/vertices to the left, right, top, and bottom; the eight-connected graph adds the diagonal neighbors.
For both, vertices on the image boundary omit edges in the direction of the boundary.

Now, we can formalize a \textbf{high-dimensional image} as a discrete function 
${\imgHighDim : \vertices \to \mathbb{R}^\numAttrDimensions}$ 
from the spatial domain of the image~\imgGraph\ to the attribute space~
${\mathbb{R}^\numAttrDimensions}$.
The number of attributes~\numAttrDimensions\ is equivalent to the number of image channels, e.g., spectral bands of a hyperspectral image.
Specifically, a pixel's high-dimensional attribute~%
${\imgHighDim(\imgPos) =\dataPointHighDim_{\imgPos}}$ 
is given by the vector  
${\dataPointHighDim_{\imgPos} = [\dataPointScalar_{\imgPos 1}, \dots, \dataPointScalar_{\imgPos \numAttrDimensions}]}$. 
In this work, we focus on 2D images but the above definitions are straightforward to extend to 3D images consisting of voxels.

Superpixels segment the image domain irregularly.
We define a superpixel segmentation~\superpixel\ as a partition of~\imgGraph\ with~$\numSuperpixel = |\superpixel|$\ disjoint components.
Each superpixel~%
$\component_p \in \superpixel$
is associated with a connected subgraph of~\imgGraph.
A \textbf{superpixel hierarchy} of $L$ levels is an ordered set of superpixel segmentations 
$\{ \superpixel^{0}, \dots, \superpixel^{L} \}$
with~$\superpixel^{0}$ containing all vertices from~\imgGraph\ as individual components and the property that each superpixel of~$\superpixel^{l+1}$ is obtained by merging one or more superpixels from~$\superpixel^{l}$.
Merging might not always be possible, see the merging criterion in \autoref{sec:method}.
We read~$\component^{l}_p$\ as the $p$-th superpixel on level $l$. 
Each superpixel in this hierarchy can be seen as a set of superpixels from a lower level, down to image pixels in the lowest level.
The number of image pixels contained in a superpixel is notated as $|\component^{l}_p|$.

In \autoref{sec:method:knn}, we introduce a weighted \textbf{neighborhood graph} $\graph = (\vertices, \edgesG)$ that represents the attribute space. %
\graph\ and~\imgGraph\ share the same vertices~\vertices, but while connectivity in~\imgGraph\ represents the spatial layout, (weighted) connectivity in~\graph\ is defined by attribute similarity, leading to a distinct set of edges~\edgesG, see \autoref{fig:notation:graphs}.
Edges in~\graph\ connect each node~$\vertex$ to the~$k$ most similar vertices in attribute space~${\mathbb{R}^\numAttrDimensions}$, defining a neighborhood~$\mathcal{N}(\ID)$.%

\section{Manifold-preserving Superpixel Hierarchy and Embedding} \label{sec:method}

We propose a novel superpixel hierarchy for use in hierarchical DR that, unlike existing superpixel methods, takes the manifold structure into account when merging superpixels.
At the same time, the superpixel approach preserves spatial coherence throughout the hierarchical aggregation.
We achieve this in the following way:
First, to merge superpixels, we need a distance measure in attribute space that respects the underlying manifold.
Following previous work on manifold preserving DR~\cite{Tenenbaum2000, Roweis2000}, we construct a neighborhood graph in attribute space as a proxy for the underlying manifold (\autoref{sec:method:knn}).
Second, we build a superpixel hierarchy that merges (super)pixels using a random-walk based distance as merging criteria (\autoref{sec:method:ImgHierarchy}).
Finally, in \autoref{sec:method:ComputeEmbeddings}, we explain how that same distance is also used as pairwise distance between superpixels in neighborhood-preserving DR such as t-SNE~\cite{VanderMaaten2008} and UMAP~\cite{McInnes2018UMAP}.

An implementation of our method as a standalone library and an interactive tool with coordinated views between image and embedding representation in the ManiVault framework~\cite{vkt2023manivault} are available at 
\href{https://github.com/alxvth/SPH/}{github.com/alxvth/SPH}
and
\href{https://github.com/alxvth/SPH_plugin/}{github.com/alxvth/SPH\_plugin}
respectively
.

\subsection{Constructing the neighborhood graph} \label{sec:method:knn}

We propose adopting a distance measure that captures the intrinsic geometric properties of the data, instead of globally applying distance metrics like Euclidean distances. 
This allows to merge superpixels and embed them while preserving the underlying manifold structure of the high-dimensional attribute space.

$k$-nearest neighbor~($k$NN) graphs are commonly used for capturing the manifold structure in high-dimensional data~\cite{Tenenbaum2000, Roweis2000}. %
An example of a $k$NN graph is given in \autoref{fig:notation:graphs}, right.
As is common in $k$NN-graph building, it is assumed that locally (i.e., given a sufficiently small $k \ll n$) the manifold can be approximated by the Euclidean distance in attribute space ${\delta(\ID, \ID[j]) = || \dataPointHighDim_{\ID} - \dataPointHighDim_{\ID[j]} ||_2^2}$.
As a result, the geodesic distances between any two data points can now be estimated as the shortest path on the graph representing the high-dimensional manifold structure~\cite[Figure~3]{Tenenbaum2000}.
To ensure that we can estimate the geodesic distance between any two points in the complete graph, we need to make sure that the graph is connected, i.e., consist of a single connected component.
To do so, we symmetrize the directed $k$NN-graph and connect disconnected components using a minimum spanning tree approach.
The full scheme is laid out in Appendix~\ref*{supp:symKnn}.

Once the symmetrized and connected graph~\graph\ has been built, we could directly estimate the geodesic distance between two points on the manifold as the length of the shortest-path between their corresponding vertices in~\graph. %
We provide more detail on this approach in Appendix~\ref*{supp:geodesic}.
However, calculating this for arbitrary pairs of nodes, as needed for the neighborhood embedding, is computationally prohibitive.
Further, it has been shown that shortest paths can over-represent shortcuts and thus misrepresent the manifold~\cite{leeNonlinearDimensionalityReduction2005, Lafon2006}.
Instead, inspired by the diffusion distance used in Diffusion Maps~\cite{Lafon2006} as well as previous hierarchical DR methods~\cite{Pezzotti16HSNE, Marcilio-Jr2021HUMAP}, we propose to use a robust similarity measure based on features obtained from random walks on~\graph.
We discuss this approach in combination with the hierarchy construction and embedding process in the following sections.

\subsection{Constructing the image hierarchy} \label{sec:method:ImgHierarchy}

To construct the desired manifold preserving superpixel hierarchy, we first have to define a proper similarity measure to govern the merging of pixels and later use it to define neighborhoods in attribute space for similarity embeddings.
We use random walks on graph~\graph\ to compute a distance that integrates over all paths between points, rather than relying solely on the shortest path.
We will use the same distance first to compare individual pixels and later superpixels.
Note, that while it might seem wasteful to calculate these distances between non-neighboring superpixels in the image space purely for the purpose of merging, we reuse these distances between arbitrary superpixels in the embedding process (see \autoref{sec:method:ComputeEmbeddings}).

We build the hierarchy bottom up, starting on the data level~${l=0}$, using the attribute space graph~\graph\ discussed in \autoref{sec:method:knn}.
For each vertex in~\graph, we start~\randomWalkNumber\ walks with~\randomWalkSteps\ steps governed by the edge weights of~\graph.
On the data level, we use a Gaussian kernel to transform the edge weight, i.e., the distance between nodes $\delta(\ID, \ID[j])$ in attribute space, into transition probabilities %
\begin{equation} \label{eq:probDistOnData}
	p^0_{\ID[j] | \ID} = \frac{\exp\left( - \delta(\ID, \ID[j]) / \sigma_{\ID} \right)}{ \sum_k \exp\left( - \delta(\ID, \ID[k]) / \sigma_{\ID} \right)} \quad \textrm{with}\ \ID[j], \ID[k] \in \mathcal{N}(\ID)
\end{equation}
following the formulation of t-SNE~\cite{VanderMaaten2014tSNE}, where $\sigma_{\ID}$ is calculated for a user specified perplexity~$u$.
Further, $u$ is directly coupled to the number of neighbors $k$ for the original $k$NN graph~\graph\ as ${k = |\mathcal{N}(\ID)| = 3 * u}$~\cite{VanderMaaten2014tSNE}.
We follow Kobak et al.~\cite{Kobak2019} and limit the perplexity~$u$ to the range $\left[ 10, 100 \right]$.
The resulting transition probabilities will be high for small attribute distances and low for large distances.
Notably, a random walk is confined to the edges of~\graph\ in attribute space.
The self-step probability $p^0_{\ID | \ID}$ is zero as~\graph\ does not contain self loops.
However, a walk may visit any node multiple times and thus also return to its starting vertex.

The random walks populate a sparse matrix~\transitions. 
A random walk starting at vertex~\ID\ in the attribute graph~\graph\ populates the matrix row~$\transitions(\ID, -)$. %
After each step to a vertex $\vertex[\textbf{j}]$ we increase the feature~$\transitions(\ID, \ID[j])$ by a weight. %
To indicate decreasing importance of later steps along the walk, we define the weight of the first step of the walk as $1$ and decrease it exponentially as the walk progresses.
Finally, we normalize each row~$\transitions(\ID, -)$ such that all entries sum to 1.
We can interpret this row then as a probability distribution, dictating the probability that any vertex $\vertex[\textbf{j}]$ would be in the neighborhood of $\vertex$.
\autoref{fig:Method} illustrates the coverage of the random walks, i.e., the vertex feature, for two vertices in green and blue.

We use the transition probability vector~$\transitions(\ID, -)$ as a descriptor (feature) of the high-dimensional neighborhood of vertex~\ID\ in the attribute graph~\graph. 
Now, we can define the similarity between two vertices, and later superpixels, by the Bhattacharyya coefficient~$BC$ which measures the overlap of two probability distributions and is straightforward to interpret, in our case:
\begin{equation} \label{eq:simRandomWalks}
	BC^{l}(\component^{l}_r, \component^{l}_s) = \sum_{\component^{l}_t \in \superpixel^{l}} \sqrt{ \transitions^{l}(\component^{l}_r, \component^{l}_t)\ \transitions^{l}(\component^{l}_s, \component^{l}_t) }
\end{equation}
The Bhattacharyya coefficient always lies within $[0, 1]$ where 0 indicates no overlap/similarity and 1 identical distributions.
Note that on the data level~$l=0$, a superpixel~$\component^{0}_i$\ corresponds to a pixel at ${\ID = (\posx, \posy)_{\id}}$, yielding
\begin{equation} \label{eq:bc0}
    BC^{0}(\ID, \ID[j]) = \sum_{\ID[k] \in \superpixel^{0}} \sqrt{ \transitions^{0}(\ID, \ID[k])\ \transitions^{0}(\ID[j], \ID[k]) }.
\end{equation}

With the similarity measure in place, we can now merge superpixels and build the hierarchy.
We largely adapt the Bor\r{u}vka superpixel hierarchy algorithm~\cite{Wei2018SuperpixelHierarchy} with the Bhattacharyya coefficient as similarity measure.
For each superpixel, we compare all spatial neighbors, i.e., connected vertices in~\imgGraph, and merge the vertex with the highest Bhattacharyya coefficient, i.e., the most similar one according to the transition probability vector.
Pixels are either 4- or 8-connected in~\imgGraph, but on higher abstraction levels superpixel neighbor-relations in~\imgGraph\ are usually less regular.
The random walks used to define $BC$ (\autoref{eq:simRandomWalks}) have a limited length~\randomWalkSteps\ and thus it is likely that $BC$ is $0$ for many pairs of dissimilar pixels.
As a result, in contrast to the global distance measures presented for color images~\cite{Wei2018SuperpixelHierarchy} or explicitly calculating shortest paths in~\graph, all neighbors of a vertex in~\imgGraph\ may have the same~$BC$ of~$0$.
To address this situation, we deviate from the original algorithm and purposely do not merge this superpixel with any neighbor on this level of the hierarchy.
However, as other superpixels grow on higher hierarchy levels, a new link may add a non-zero similarity leading to a merger.

Merging superpixels gives us the structure for the image graph~\imgGraph\ for the next hierarchy level~${l+1}$.
We still need similarities between the new superpixels that respect the structure of the original attribute-based graph~\graph.
Since both graphs share the same vertices, conceptually, we can achieve this by simply merging the same nodes in~\graph\ and going through the same process described above to calculate the random walks, followed by transition probability distributions and finally the $BC$.
We can achieve a similar result, without starting new random walks, by directly merging the rows and columns of the transition matrix~$\transitions^{l}$ following the merging pattern of~\imgGraph.
We add and re-normalize the transition probabilities when merging rows and columns resulting in the new transition matrix~$\transitions^{l+1}$.
\autoref{fig:Method}~(3-5) illustrates the vertex and feature merging.
The merged features of the meta vertices now describe the local graph structure of all pixels contained in the corresponding superpixel.

\subsection{Computing embeddings on each abstraction level} \label{sec:method:ComputeEmbeddings}

To embed the data, we generally follow the process of existing non-linear DR methods like t-SNE or UMAP, and only replace their calculation of the high-dimensional conditional probability matrix $P$~\cite[Equation 1]{VanderMaaten2014tSNE}.
On the pixel level~${l=0}$, we can directly use the original t-SNE probabilities as defined in \autoref{eq:probDistOnData}. %
On higher abstraction levels~${l>0}$, we simply replace the distance $\delta$ in \autoref{eq:probDistOnData} with the Bhattacharyya distance~$d_{Bhat} = -\ln{BC}$, leading to
\begin{equation} \label{eq:transitionLevel}
	p_{\component_s | \component_r} = \frac{\exp\left( - d_{Bhat}(\component_r, \component_s) / \sigma_{\component_s} \right)}{ \sum_k \exp\left( - d_{Bhat}(\component_r, \component_k) / \sigma_{\component_s} \right)}  \quad \textrm{with}\ \component_s, \component_k \in \mathcal{N}(\component_r),
\end{equation}
where $\component_r$ and $\component_s$ refer to superpixels on the corresponding level. 
$\sigma_{\component_s}$ is determined as for \autoref{eq:probDistOnData} based on level-specific perplexity~${u = \max\left(10,\ \min(|\superpixel^{l}| / 100, \ 100) \right)}$ where $|\superpixel^{l}|$ is the number of superpixels on the level, again following Kobak et al.~\cite{Kobak2019}.

The Bhattacharyya coefficient~$BC$ in~$d_{Bhat}(\component_r, \component_s)$
is equivalent to the dot product of the two corresponding rows in~\transitions\ after taking the square-root of each element (cf. \autoref{supp:derivations}). 
Further,~\transitions\ is a very sparse matrix which allows for efficient computation of a dissimilarity matrix $D = \left[ d_{Bhat}(\component_r, \component_s) \right]$ on each level, instead of computing each entry individually, as:
\begin{equation} \label{eq:probDistOnLevel}
	D = - \ln \left( \sqrt{\transitions}\ \sqrt{\transitions}^\top \right) 
\end{equation}
where $\sqrt{\transitions} = \left[\sqrt{ \transitions(\component_r, \component_t)} \right]$ is the element-wise square root of~\transitions. %
$D$~is inherently symmetric since the Bhattacharyya coefficient is symmetric, reducing the computational complexity further.
We also symmetrize the conditional probabilities~$P = \left\{ p_{\ID \ID[j]} \right\}$ with~$p_{\ID \ID[j]} = ( p_{\ID | \ID[j]} + p_{\ID[j] |\ID} ) / 2$ as in t-SNE~\cite{VanderMaaten2014tSNE}.

\begin{figure*}[t]
    \centering
    \includegraphics[width=\linewidth]{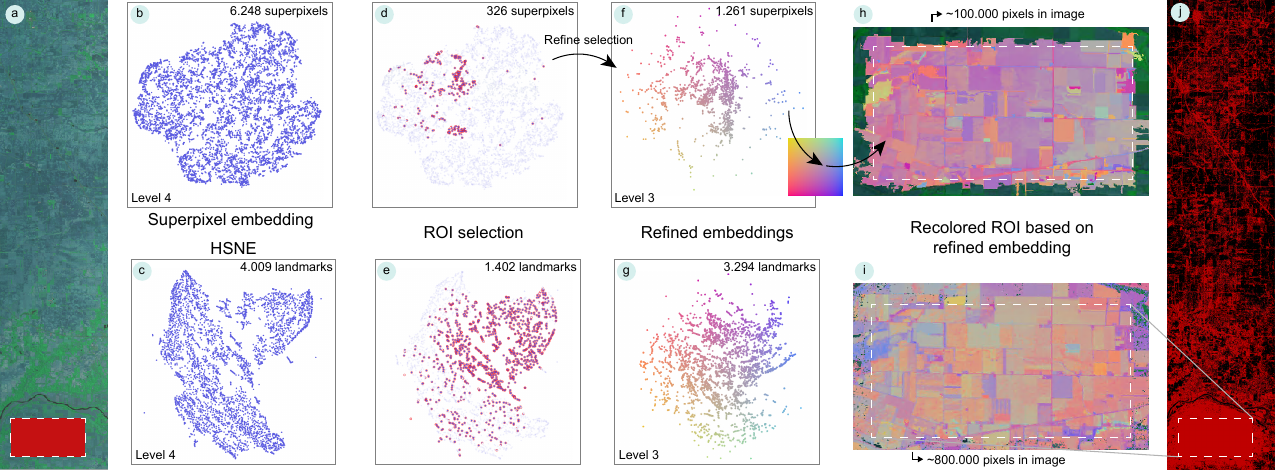}
    {\phantomsubcaption\label{fig:IndianPinesLargeExploration:a}}%
    {\phantomsubcaption\label{fig:IndianPinesLargeExploration:b}}%
    {\phantomsubcaption\label{fig:IndianPinesLargeExploration:c}}%
    {\phantomsubcaption\label{fig:IndianPinesLargeExploration:d}}%
    {\phantomsubcaption\label{fig:IndianPinesLargeExploration:e}}%
    {\phantomsubcaption\label{fig:IndianPinesLargeExploration:f}}%
    {\phantomsubcaption\label{fig:IndianPinesLargeExploration:g}}%
    {\phantomsubcaption\label{fig:IndianPinesLargeExploration:h}}%
    {\phantomsubcaption\label{fig:IndianPinesLargeExploration:i}}%
    {\phantomsubcaption\label{fig:IndianPinesLargeExploration:k}}%
    \vspace{-3mm}
    \caption{\textbf{Indian Pines Exploration:} %
    (a) a false color image based on data channels 20 (587 nm, red), 76 (1090 nm, green) and 130 (1591 nm, blue) with a ROI marked in red.
    (b) and (c) show the 4th abstraction level embedding of our superpixel embedding and HSNE, respectively;
    (d) and (e) highlight the superpixels and landmarks which correspond to the ROI.
    (f) and (g) show refined embeddings of the highlighted subsets on a lower abstraction level.
    (h) and (i) recolor a cutout of the image based on the refined embeddings using an overlaid 2D colormap
    Finally, (j) indicates the pixels in the full images which are represented by the HSNE refinement, whereas our superpixel refinement extends only slightly around the ROI.
    }%
    \label{fig:IndianPinesLargeExploration}%
    \vspace{-3mm}
\end{figure*}

\paragraph*{Embedding with UMAP.}
Above, we used t-SNE as basis for defining the joint probabilities~$p_{\ID \ID[j]}$ but neither the hierarchy creation nor embedding layout are limited to this choice. 
For example, we can adopt the probability definition from UMAP as:
\begin{equation} \label{eq:probDistUMAP}
	p^0_{\ID[j] | \ID} = \exp \left( - \frac{ \delta(\ID, \ID[j]) - \rho_{\ID} }{ \sigma_{\ID} } \right) \quad \textrm{with}\ \ID[j] \in \mathcal{N}(\ID)
\end{equation}
where $\rho_{\ID}$ is the distance of \ID\ to its closest neighbor and $\sigma_{\ID}$ is set such that $\sum_{\ID} p_{\ID[j] | \ID} = \log_2 k$ as in the original UMAP formulation~\cite{McInnes2018UMAP}.
On the abstraction levels, the probability distributions $p_{\component_s | \component_r}$~are defined analogously to \autoref{eq:probDistOnLevel} using the Bhattacharyya distance.
Here, we perform the conditional probability symmetrization with ~$p_{\ID \ID[j]} = p_{\ID | \ID[j]} + p_{\ID[j] |\ID} - p_{\ID | \ID[j]} p_{\ID[j] |\ID}$. 
The rest of the described algorithm remains unchanged.
Supplemental Figures~\ref*{fig:IndianPinesSmall:a} and~\ref*{fig:IndianPinesSmall:b} show multiple abstraction levels and embeddings for the Indian Pines dataset~\cite{IndianPinesData} with t-SNE and UMAP probabilities, respectively.

\subsection{Subset embedding refinement} \label{sec:method:refine}
With the above, we can now embed each level of the hierarchy as a whole.
For effective explorative analysis, we need to allow gradual zooming into more and more refined subsets of the data on lower embedding levels, similar to existing hierarchical DR methods~\cite{Pezzotti16HSNE, Marcilio-Jr2021HUMAP}.
Such a refinement operation starts with a selection of superpixels~$\widetilde{\superpixel}^{\,l}$ on level~$l$, e.g., by manual selection in the embedding or image.
We can then retrieve the more detailed set of superpixels~$\widetilde{\superpixel}^{\,l-1}$ of which the superpixels in~$\widetilde{\superpixel}^{\,l}$ are composed of.
To embed this subset, we slice the complete probability matrix~$P^{\,l-1}$ of level~${l-1}$ described in \autoref{sec:method:ComputeEmbeddings} to receive the sub-matrix~$\widetilde{P}^{\,l-1}$ consisting of only the rows and columns corresponding to~$\widetilde{\superpixel}^{\,l-1}$.
Finally, we normalize the rows of~$\widetilde{P}^{\,l-1}$ to represent a probability distribution.

Superpixel selections in the image space can contain superpixels without any connections in the transition matrix. 
t-SNE will not be able to place these superpixels close to any other, leading to isolated points occupying large peripheral parts of the embedding space (e.g., \autoref{fig:IndianPinesLargeExploration:f}).
While this can be useful to identify outliers in the current region of interest, it also compresses the actual structure to the center of the embedding.
If desired, we can relax the selection criteria of lower-level superpixels in order to introduce additional, non-selected superpixels from level~$l-1$. %
Using a user-defined threshold~$\gamma \in [0, 1]$, any superpixel~$\component^{l-1}_j$ that is connected to a selected~$\widetilde{\superpixel}^{\,l-1}_i$ with~$p^{l-1}_{\ID \ID[j]} > \gamma$ is also added to the refined superpixels. 
This reduces the amount of isolated superpixels while keeping the data structure (e.g., see \autoref{fig:IndianPinesNonExactRefinement}).

\section{Validation} \label{sec:eval}

In this section, we first present two use cases with real-world data to illustrate the application of our method and, secondly, evaluate the superpixel hierarchy quantitatively. 

\subsection{Use Case: Exploring Hyperspectral Images} \label{sec:eval:hyper}

Hyperspectral images contain information about a large spectrum of light, in contrast to three color channels in RGB images.
Here, we present an example based on the Indian Pines Test Site~3~\cite{IndianPinesData} dataset. 
The image depicts fields (e.g., corn and soy), forests, roads, rivers and houses from an aerial perspective. 
It measures ${614 \times 2,678 \approx 1.6 \textrm{M}}$ pixels with~200 channels.
The pixel resolution is roughly \SI{20}{\meter}~$\times$~\SI{20}{\meter} and the channels contain electromagnetic spectral information from \SI{400}{\nano\meter} to \SI{2400}{\nano\meter} sampled at~10\si{\nano\meter}.
We exclude~$20$ of the original~$220$ channels due to their low information, as suggested by Gualtieri and Cromp~\cite{gualtieri1999}. %

We compare our superpixel embedding exploration with an image-coupled HSNE exploration~\cite{Vieth23} which uses a conventional image-agnostic hierarchical data representation, and is representative of the state-of-the-art hierarchical embedding methods for data exploration.
Our superpixel hierarchy is computed with~$\randomWalkNumber = 50$ random walks with~$\randomWalkSteps = 25 $\ steps each.
We empirically found $\randomWalkNumber \in [20, 50]$ and $\randomWalkSteps \in [10, 50]$ to yield reliable results. 
The HSNE hierarchy is computed with 4 levels using default settings.

Exploring large hyperspectral images typically involves the identification and analysis of interesting spatial-spectral regions, i.e., a combination of high-dimensional attribute and image layout characteristics.
We start our exploration on hierarchy level 4, shown in 
\Autoref{fig:IndianPinesLargeExploration:b, fig:IndianPinesLargeExploration:c} for our superpixel hierarchy and HSNE, respectively, as either contain a manageable number of superpixels/landmarks. %
The ${\smallsim} 100,000$ pixel ROI is covered by $326$ of the ${\smallsim} 6,000$ superpixels on the fourth abstraction level of our hierarchy and $1,402$ of ${\smallsim} 4,000$ landmarks in the respective HSNE level, highlighted in \Autoref{fig:IndianPinesLargeExploration:d, fig:IndianPinesLargeExploration:e}.
As the HSNE hierarchy has no notion of the spatial layout, many more landmarks than in our hierarchy are needed to represent the same region on a comparable level of detail.

In this example, we focus on a ROI in the lower part of the image as indicated in \autoref{fig:IndianPinesLargeExploration:a}.
Focusing on the ROI, we refine the selected embedding points to explore data similarities on a lower level of abstraction.
Note, that for presentation here, we connect the embedding with the image representation by a recoloring approach, mapping 2D embedding positions to color in the image with a 2D colormap by Bernard et al.~\cite{Bernard2015}.
In our interactive implementation further detail can also be gathered through interactive linked selections.
The $326$ superpixels from level 4 expand to $1,261$ superpixels on level 3 (\autoref{fig:IndianPinesLargeExploration:f}), still fewer than needed to represent the ROI on level 4 of the HSNE hierarchy, leaving room for further increasing the detail by zooming into level 2 of the superpixel hierarchy. %
HSNE expands to $3,294$ refined landmarks on level 3 (\autoref{fig:IndianPinesLargeExploration:g}).
Again, due to the spatially-agnostic landmarks in HSNE these landmarks cover a much larger area, and with ${\smallsim} 800,000$ pixels almost half the image (see \autoref{fig:IndianPinesLargeExploration:k}).

\begin{figure}[b]
    \centering
    \vspace{-2mm}
    \includegraphics[width=\linewidth]{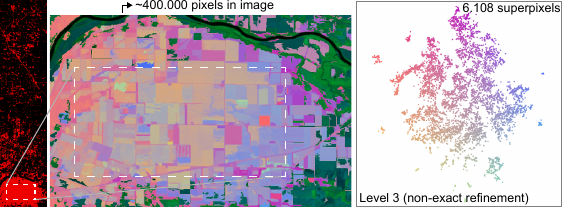}
    \vspace{-2mm}
    \caption{\textbf{Non-exact refinement} %
    as discussed in \autoref{sec:method:ComputeEmbeddings}.
    Here, using~$\gamma = 0.01$ can improve the embedding structure by including additional superpixels not in the ROI (c.f. \autoref{fig:IndianPinesLargeExploration:f}).
    }%
    \label{fig:IndianPinesNonExactRefinement}%
    \vspace{-2mm}
\end{figure}

Notably, the refined superpixel embedding shows better distinguishable clusters in the center, where the bulk of embedded superpixels are located, but also shows more isolated superpixels in its periphery than the ROI-refined HSNE embedding due to individual disconnected superpixels (see \Autoref{sec:method:refine}). 
As discussed in \Autoref{sec:method:refine}, a possible means to improve connectivity of these disconnected superpixels is to perform a non-exact refinement.
In \autoref{fig:IndianPinesNonExactRefinement} we included additional superpixels by setting ~$\gamma = 0.01$. 
The number of superpixels grows five-fold to $6,108$, and nearly twice the size of the level 3 HSNE embedding, but now shows clear structure, both in embedding and image space.
In the embedding in \autoref{fig:IndianPinesNonExactRefinement}, clusters are clearly identifiable and more separated than in the HSNE embedding in \autoref{fig:IndianPinesLargeExploration:g}.
In image space, borders of relevant image regions like fields, roads and rivers are clearly delineated.

In summary, our superpixel hierarchy keeps selections in image space more concise in the hierarchy and thus allows for more structure in and faster to compute embeddings with fewer data points embedded at the same level of detail compared to HSNE.

\subsection{Use Case: Exploring CyCIF Images} \label{sec:eval:cycif}

\begin{figure}[b!]
    \centering
    \vspace{-2mm}
    \includegraphics[width=\linewidth]{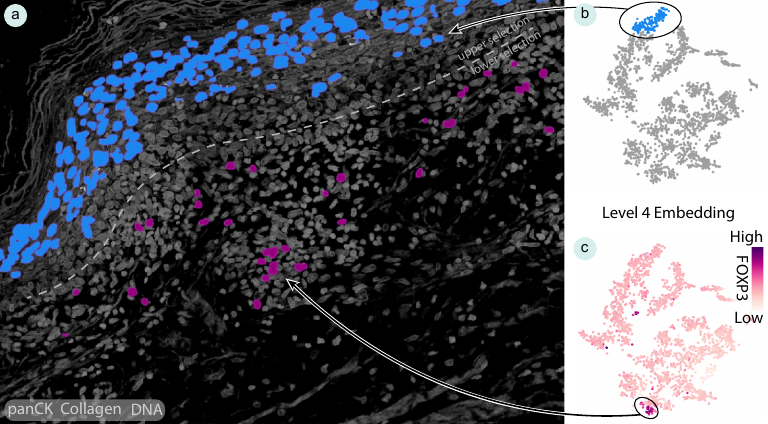}
    {\phantomsubcaption\label{fig:cicyf:overview:a}}%
    {\phantomsubcaption\label{fig:cicyf:overview:b}}%
    {\phantomsubcaption\label{fig:cicyf:overview:c}}%
    \vspace{-2mm}
    \caption{\textbf{CyCIF Data.} %
    (a) Grayscale image based on two structural markers and DNA with two selections highlighted, corresponding to two clusters in the level 4 superpixel embedding (b, c).
    }%
    \label{fig:cicyf:overview}%
\end{figure}

Analyzing the function of and interplay between cells in tissue is of major interest in systems biology, e.g., for researching cancer.
Various high-dimensional imaging methods are used in this domain. %
One of them is Cyclic Immunofluorescence (CyCIF)~\cite{cycif}, measuring the abundance of proteins in tissue with a spatial resolution in the micrometer range.
CyCIF creates images with tens of channels, each representing one selected protein.
Here, we use a CyCIF dataset of cancerous skin tissue~\cite{Yapp2023}, also used in the IEEE VIS 2025 Bio+MedVis Challenge~\cite{BioMedVisChallenge2025}, containing $10,908 \times 5,508 \times 194$ voxels and $54$ channels in a multi-resolution representation. %
For illustration, we use a cutout (see Supplemental Figure~\ref*{fig:cicyf:largeImage}) sized $2,000 \times 1,500$ pixels at resolution level 4 (\SI{1.12}{\micro\meter} pixel size).
As suggested by the original authors (Appendix~\ref*{supp:cicyf}), we select $27$ channels and log-normalize the values.
\autoref{fig:cicyf:overview:a} illustrates the cutout by mapping three channels, indicative of tissue structure and DNA (indicating cell nuclei), to grayscale.

\begin{figure*}[ht!]
    \centering
    \vspace{0mm}
    \includegraphics[width=\linewidth]{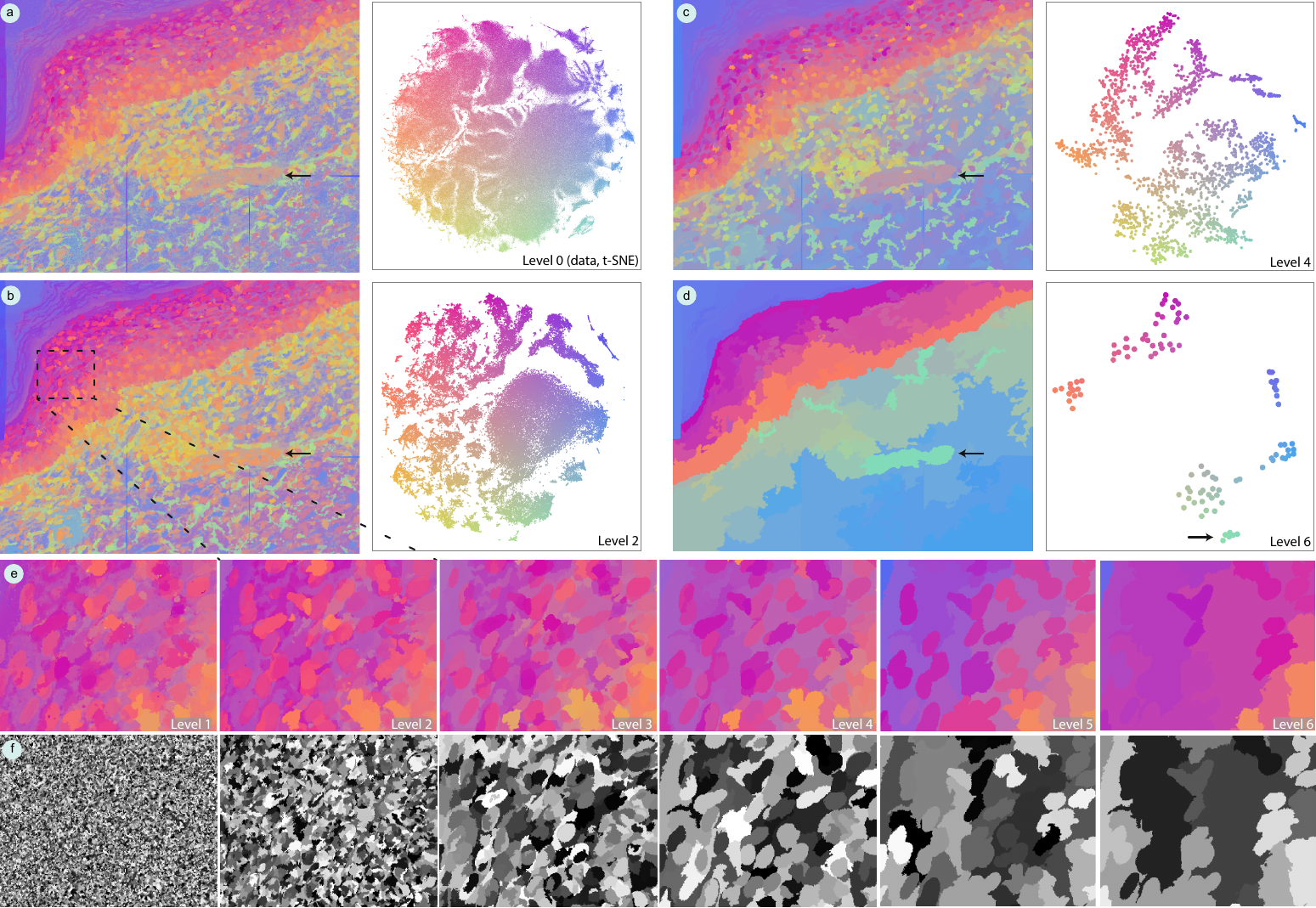}
    {\phantomsubcaption\label{fig:cicyf:hierarchy:a}}%
    {\phantomsubcaption\label{fig:cicyf:hierarchy:b}}%
    {\phantomsubcaption\label{fig:cicyf:hierarchy:c}}%
    {\phantomsubcaption\label{fig:cicyf:hierarchy:d}}%
    {\phantomsubcaption\label{fig:cicyf:hierarchy:e}}%
    {\phantomsubcaption\label{fig:cicyf:hierarchy:f}}%
    \vspace{-4mm}
    \caption{\textbf{Hierarchical Embeddings on CyCIF data:} %
    (a-d) Embeddings on several levels of abstraction and corresponding recolored images, using the colormapping discussed in \autoref{sec:eval:hyper}.
    (e, f) Zoom in on a ROI as highlighted in (b), recolored according to the embedding and with random gray values to better distinguish between superpixels, respectively.
    }%
    \label{fig:cicyf:hierarchy}%
    \vspace{-4mm}
\end{figure*}

The profile of protein abundance is used to characterize cell types, and the spatial distribution of cells provides cues on interactions between cells, and between cells and other tissue structures. 
The segmentation of cells and the subsequent analysis of their spatial neighborhood are key concerns in current research.
Our superpixel hierarchy and embedding can be helpful in combining segmentation and exploration into a single workflow.
To illustrate this, we compute our superpixel hierarchy for the described cutout using~$\randomWalkNumber = 30$,~$\randomWalkSteps = 10$ and~$k = 90$.
We use fewer random walks for this larger image compared to \autoref{sec:eval:hyper} to reduce the computational and memory load.
\autoref{fig:cicyf:overview:b} shows the level~$4$ embedding with $4,104$ superpixels.
Here, it can already be seen that many superpixels on this level partially or fully match the outline of cells in the image (see e.g., the exemplative selections in \autoref{fig:cicyf:overview:a} showing many compact oval structures common for immune cells.)

A straightforward step to identify specific cells is to explore individual channels, corresponding to proteins indicative of corresponding cell types.
As an example, we show the average abundance of the \texttt{FOXP3} protein, indicative of regulatory T-cells, per superpixel using colormapping on the level 4 embedding in \autoref{fig:cicyf:overview:c}.
One cluster in the embedding stands out with all superpixels showing bright pink color, indicating high abundance of \texttt{FOXP3}.
The superpixels correspond to multiple cells scattered along the diagonal of the image, indicated in \autoref{fig:cicyf:overview:a}.
Tumor infiltration by regulatory T-cells can be an important indicator for survival in various cancer types~\cite{Pan2025Tregs}.
Straight forward identification and localization of such cells as shown in this simple example can be of great help in related analysis tasks. 

\autoref{fig:cicyf:hierarchy} shows more detail on the created hierarchy, with embeddings and recolored images following the scheme described above for hierarchy levels 0, 2, 4, and 6.
We follow Kobak et al.~\cite{Kobak2019} and initialize the embeddings with the first two PCA components rather than randomly.
This helps preserving global structure in SNE-based approaches, and thus improves comparability between levels.
As can be seen, the different levels of our hierarchy represent different aspects of the data.
On the high abstraction level 6 (\autoref{fig:cicyf:hierarchy:d}), the embedding splits the tissue in two main regions (besides the blue/purple colored background), colored in green (bottom of embedding) and orange/pink (top of embedding).
This high-level separation is consistent with the dermal-epidermal junction (DEJ) described by Yapp et al. for this dataset~\cite[Figure 5]{Yapp2023}.

When requesting more detail from the hierarchy on subsequent levels (Figures \ref{fig:cicyf:hierarchy:a}--\ref{fig:cicyf:hierarchy:c}), the round-ish immune cells become clearly visible, with the different cell types in dermis and epidermis still creating a clear boundary.
More detail on the merge sequence can be seen in \Autoref{fig:cicyf:hierarchy:e, fig:cicyf:hierarchy:f}, showing a cutout highlighted in \autoref{fig:cicyf:hierarchy:b}.
\autoref{fig:cicyf:hierarchy:e} uses the same recoloring scheme as above.
To better distinguish individual superpixels, \autoref{fig:cicyf:hierarchy:f} assigns random gray values to each.
It can be seen that around level 4 superpixels largely delineate individual cells, while below these further subdivide into still similar components (as indicated by the similar color in \autoref{fig:cicyf:hierarchy:e}).

An interesting detail are the blood vessels that already on level 6 create a separate cluster (mint green \autoref{fig:cicyf:hierarchy:d}).
While the cluster moves towards the center of the embedding on more detailed hierarchy levels, creating a grey-orange color, especially the vessel also analyzed by Yapp et al.~\cite[Figure 2a]{Yapp2023} stays clearly visible (arrows in Figures \ref{fig:cicyf:hierarchy:a}--\ref{fig:cicyf:hierarchy:d}).
Additional structure within the vessel is also clearly visible on hierarchy level 2 (\ref{fig:cicyf:hierarchy:b}), with several bright orange spots offset from the more grey-ish vessel background.

An in-depth analysis of the segmentation quality of the superpixel hierarchy goes beyond the scope of this work.
However, these initial findings are a promising indication of a potential joint segmentation and exploration of single-cell data, which are typically two completely separate workflows.

\subsection{Quantitative Hierarchy Evaluation} \label{sec:eval:quant}
Here, we compare our superpixel hierarchy to existing superpixel hierarchy methods that are applicable to high-dimensional images, i.e., methods that do not exclusively rely on color-space distances or project data to three channels in an initial step.
It should be noted that the goal of our work is to produce hierarchical similarity embeddings for image exploration, rather than purely a superpixel segmentation.
Since none of the existing methods are suitable for this goal, we aim for comparable quality and not strictly outperforming others.
Nonetheless, we focus on the hierarchy quality here, and do not explicitly evaluate the embedding layout, as we do not modify the embedding algorithms after defining high-dimensional transition probabilities in \Autoref{eq:probDistOnLevel, eq:probDistUMAP}, and use the same gradient descent optimization as t-SNE and UMAP respectively. 

We compare to four other methods, specifically, FH~\cite{Felzenszwalb2004}, Entropy Rate Superpixel (ERS)~\cite{Liu2011Entropy}, SLIC~\cite{achantaSLIC2012} (without feature conversion to Lab colorspace) and a SLIC variant modified for hyperspectral images (BB)~\cite{barbato22}.
Additionally, we compare our random-walk based spatial hierarchy method (SPH) to the geodesic distance variation discussed in \autoref{sec:method:problem} (SPH geodesic). 

For comparison, we use two hyperspectral image data sets with a ground truth segmentation. %
The Indian Pines data set~\cite{IndianPinesData} is a labelled subset of the dataset described in \autoref{sec:eval:hyper}, comprising~${145 \times 145 \approx 21,000}$ pixels with $200$ channels and contains 16 labels (exclusive background).
Only this region of the entire image contains segmentation labels.
The Salinas data set~\cite{SalinasData} comprises~${512 \times 217 \approx 111,000}$ pixels with $204$ channels and 16 labels (exclusive background).
We pre-processed both images by clipping all pixel intensities at the 98th global percentile and subsequently normalizing each channel to unit maximum.

Stutz et al.~\cite{stutzSuperpixelsEvaluationStateoftheart2018} give a comprehensive overview of commonly used evaluation metrics for superpixel algorithms.
We measured Undersegmentation Error (UE) and Explained variation (EV). 
UE measures how much superpixels extend outside an overlapping ground truth segment. 
If all superpixels completely lie within ground truth segments, the UE is zero.
Consequently, larger superpixels on higher abstraction levels will produce increasing UE.
EV measures how well superpixels represent the image %
by comparing superpixel means with global means.
Both measures strongly depend on the number of superpixels in a segmentation.
Since we cannot force all methods to produce the same number of superpixels, we compute the UE and EV over range of segmentations and summarize a method's performance following~\cite{stutzSuperpixelsEvaluationStateoftheart2018} as the normalized area under the resulting UE and (1 - EV) curves, AUE and AEV.
For fair comparison, a grid search was performed to identify optimal parameter settings for each method, and only the best-performing (lowest AUE) configurations are reported.

\begin{table}[t]
\centering
\caption{\textbf{Explained Variation and Undersegmentation Error:} normalized area under the UE and (1 - EV) value curves in $\%$, lower is better for both AUE and AEV.}
\label{tab:quant_auc}
\vspace{-1mm}
\begin{tabular}{lS[table-format=2.2]S[table-format=2.2]S[table-format=2.2]S[table-format=2.2]}
    \toprule
    & \multicolumn{2}{c}{Pines {\small\cite{IndianPinesData}}} & \multicolumn{2}{c}{Salinas {\small\cite{SalinasData}}} \\
    \midrule
    \textbf{Method} & {\textbf{AUE}} & {\textbf{AEV}} & {\textbf{AUE}} & {\textbf{AEV}} \\
    \midrule
    FH {\small\cite{Felzenszwalb2004}} & \textbf{11.17} & {10.45}  & \zerospace \textbf{5.66} & {1.75} \\ %
    ERS {\small\cite{Liu2011Entropy}} & {15.46} & {12.78} & \zerospace {8.46} & {3.93} \\ %
    SLIC {\small\cite{achantaSLIC2012}} & {13.72} & {13.11} & \zerospace {5.16} & {4.15} \\ %
    BB {\small\cite{barbato22}} & {16.25} & \zerospace {9.15} & {10.27} & {4.53} \\ %
    SPH geodesic {\small\textit{(ours)}} & {13.06} & \zerospace {8.92} & \zerospace {7.35} & \textbf{1.56} \\
    SPH $k$NN {\small\textbf{(ours)}} & {12.47} & \zerospace \textbf{8.90} & \zerospace {7.05} & {1.70} \\
    \bottomrule
\end{tabular}
\vspace{-3mm}
\end{table}

As shown in \autoref{tab:quant_auc}, our method generally matches or outperforms the other algorithms in terms of AEV, while FH yields better results for AUE.
However, it should be noted that the area under the curves is skewed to put more emphasis on hierarchy levels with a large number of superpixels, masking FH's rather weak performance for small numbers of superpixels (Supplemental Figures~\ref*{fig:qaunt:results_pines} and ~\ref*{fig:qaunt:results_salinas}).
Appendix~\ref*{supp:quant} provides more detail on the analysis, including an alternative logarithmic scaling of AUE and AEV that weighs the exponential growth in the hierarchy more equally (see Supplemental Figures~\ref*{fig:qaunt:results_pines} and ~\ref*{fig:qaunt:results_salinas}, and Supplemental Table~\ref*{tab:quant_auc_log}).
Further, both FH and SLIC apply Gaussian smoothing to the input images to reduce noise, which the other methods do not benefit from. 
A qualitative perspective is provided in Supplemental Figures~\ref*{fig:comparison:pines} and~\ref*{fig:comparison:salinas} showing several segmentation levels for all superpixel methods for the Indian Pines and Salinas data sets, respectively.

\section{Discussion} \label{sec:discussion}

The length \randomWalkSteps\ and number \randomWalkNumber\ of random walks on the attribute-based graph \graph\ are free parameters in our method.
The ability of these random walks to create features that are representative of the high-dimensional attribute data structure is inherently tied to the data itself. 
Consequently, optimal parameter settings may vary across datasets.
We empirically found $\randomWalkSteps \in [10, 50] $ and $\randomWalkNumber \in [20, 50]$ to yield good results with all tested datasets, with smaller values requiring less compute time and memory.
Random walks on graphs have been thoroughly analyzed in prior research~\cite{Lovasz1993}, and future work might consider how these findings can inform robust data-driven choices for our random walk settings. 
Node2vec~\cite{groverNode2vec2016} and DeepWalk~\cite{perozziDeepWalk2014} leave these parameters free with a similar reasoning and default to 10 walks from each vertex with 80 and 40 steps, respectively. %
Other works use heuristic measures to set the walk length.
PHATE~\cite{Moon2019} bases this on measured entropy, while Kim et al.~\cite{kim2025} use the concept of \emph{cover time}, i.e., the number of steps it takes to visit all vertices of a graph.

As visible in the image recolorings in Supplemental Figure~\ref*{fig:IndianPinesLargeRecolored} the global position of merged superpixels in the t-SNE embedding is not stable across hierarchy levels.
This happens due to following t-SNEs default random initialization.  
An initialization using PCA as in \autoref{sec:eval:cycif} or a scheme similar to the one presented in~\cite{Vieth23}, placing points explicitly according to their parents in the hierarchy could improve the embedding coherence among levels (as applied in Supplemental Figure~\ref*{fig:IndianPinesSmall}).

We compute random walks only on the initial attribute-based graph \graph\ and essentially aggregate \graph\ on each abstraction level while merging the random walk vertex features, similar to the diffusion and coarse-graining of Multiscale PHATE~\cite{kuchroo2022}.
Other methods like HSNE~\cite{Pezzotti16HSNE} and HUMAP~\cite{Marcilio-Jr2021HUMAP} start new random walks on each new abstraction level for selecting new landmarks and defining similarities.
While the prior reduces computation, new walks on each scale provide more fine-grained control over the reproduced graph structure. 
We did experiment with new random walks, but did not observe improved results.

The superpixel hierarchy method by Wei et al.~\cite{Wei2018SuperpixelHierarchy}, which we build upon, ensures that any given superpixel is \textit{always} merged with at least one other.
In contrast, our method is allowed to \textit{not} merge a superpixel, if all Bhattacharyya coefficients $BC$ between a superpixel's feature and its spatial neighbors is~0 (see \autoref{eq:simRandomWalks}).
In fact, it is worth considering to go a step further and introduce a threshold here, ensuring that even superpixels with non-zero $BC$s are only merged if they are considered similar enough.
If the random walk length is set high, a superpixel might be merged even though their attribute vertices are located very far apart in the attribute graph \graph.
However, defining a good threshold is not trivial and would need to depend on the data structure.

\section{Conclusion} \label{sec:conclusion}

We present a superpixel hierarchy for high-dimensional images which preserves the manifold structure of the high-dimensional data.  
Each level, and subsets thereof, can be embedded based on a coherent distance measure, which enables exploring the high-dimensional attribute space.
We inform the hierarchy about the underlying high-dimensional manifold in attribute space by using random walks on an attribute-based neighborhood graph and utilize them in a modified version of Bor\r{u}vka's algorithm. 
Our image-informed hierarchy improves on previous hierarchical embedding methods with more spatially compact representations that increase embedding detail based on image space coherence.

An interesting extension of our method is the introduction of multiple levels of abstraction in a single embedding.
While such a Focus+Context approach for hierarchical embedding has been applied to exploration in attribute space~\cite{Hoellt2019FC}, utilizing it for ROIs in the image space could further aid image exploration.
Our method shows good potential for combining segmentation and exploration steps. 
Our initial showcase of single cell data exploration in \autoref{sec:eval:cycif} indicates that the superpixel cell structures could be used in subsequent cell-neighborhood analysis steps.

In summary, we have presented a hierarchical superpixel embedding method whose potential to aid the exploration of high-dimensional images is demonstrated through two use cases and quantitative analysis.

\bibliographystyle{eg-alpha-doi} 
\bibliography{references.bib}

@article{Wei2018SuperpixelHierarchy,
  title        = {{Superpixel Hierarchy}},
  author       = {Wei, Xing and Yang, Qingxiong and Gong, Yihong and Ahuja, Narendra and Yang, Ming Hsuan},
  year         = 2018,
  journal      = {IEEE Transactions on Image Processing},
  volume       = 27,
  number       = 10,
  pages        = {4838--4849},
  doi          = {10.1109/TIP.2018.2836300},
  issn         = 10577149
}

@article{Thompson2010SuperpixelEndmember,
  title        = {Superpixel {{Endmember Detection}}},
  author       = {Thompson, David R. and Mandrake, Lukas and Gilmore, Martha S. and Casta{\~n}o, Rebecca},
  year         = 2010,
  journal      = {IEEE Transactions on Geoscience and Remote Sensing},
  volume       = 48,
  number       = 11,
  pages        = {4023--4033},
  doi          = {10.1109/TGRS.2010.2070802},
  issn         = {1558-0644}
}

@article{Felzenszwalb2004,
  title        = {Efficient {{Graph-Based Image Segmentation}}},
  author       = {Felzenszwalb, Pedro F. and Huttenlocher, Daniel P.},
  year         = 2004,
  journal      = {International Journal of Computer Vision},
  volume       = 59,
  number       = 2,
  pages        = {167--181},
  doi          = {10.1023/B:VISI.0000022288.19776.77},
  issn         = {1573-1405}
}

@article{Fan2023,
  title        = {Robust {{Superpixel Segmentation}} for {{Hyperspectral-Image Restoration}}},
  author       = {Fan, Ya-Ru},
  year         = 2023,
  journal      = {Entropy},
  publisher    = {{Multidisciplinary Digital Publishing Institute}},
  volume       = 25,
  number       = 2,
  pages        = 260,
  doi          = {10.3390/e25020260},
  issn         = {1099-4300}
}

@inproceedings{Liu2011Entropy,
  title        = {Entropy Rate Superpixel Segmentation},
  author       = {Liu, Ming-Yu and Tuzel, Oncel and Ramalingam, Srikumar and Chellappa, Rama},
  year         = 2011,
  booktitle    = {Proceedings of {{CVPR}}},
  pages        = {2097--2104},
  doi          = {10.1109/CVPR.2011.5995323},
  issn         = {1063-6919}
}

@inproceedings{tangDifferentVersionsEntropy2019,
  title        = {Different {{Versions}} of {{Entropy Rate Superpixel Segmentation For Hyperspectral Image}}},
  author       = {Tang, Yiwei and Zhao, Liaoying and Ren, Lang},
  year         = 2019,
  booktitle    = {Proceedings of {{ICSIP}}},
  pages        = {1050--1054},
  doi          = {10.1109/SIPROCESS.2019.8868344}
}

@article{xuRegionalClusteringbasedSpatial2018,
  title        = {Regional Clustering-Based Spatial Preprocessing for Hyperspectral Unmixing},
  author       = {Xu, Xiang and Li, Jun and Wu, Changshan and Plaza, Antonio},
  year         = 2018,
  journal      = {Remote Sensing of Environment},
  volume       = 204,
  pages        = {333--346},
  doi          = {10.1016/j.rse.2017.10.020},
  issn         = {0034-4257}
}

@article{Yan2022HierarchicalSuperpixel,
  title        = {Hierarchical {{Superpixel Segmentation}} by {{Parallel CRTrees Labeling}}},
  author       = {Yan, Tingman and Huang, Xiaolin and Zhao, Qunfei},
  year         = 2022,
  journal      = {IEEE Transactions on Image Processing},
  volume       = 31,
  pages        = {4719--4732},
  doi          = {10.1109/TIP.2022.3187563},
  issn         = {1941-0042}
}

@article{VanderMaaten2014tSNE,
  title        = {Accelerating T-{{SNE}} Using {{Tree-Based Algorithms}}},
  author       = {{van der Maaten}, Laurens},
  year         = 2014,
  journal      = {Journal of Machine Learning Research},
  volume       = 15,
  number       = 93,
  pages        = {3221--3245},
  url          = {https://jmlr.org/papers/v15/vandermaaten14a.html}
}

@misc{McInnes2018UMAP,
  title        = {{{UMAP}}: {{Uniform Manifold Approximation}} and {{Projection}} for {{Dimension Reduction}}},
  author       = {McInnes, Leland and Healy, John and Melville, James},
  year         = 2018,
  doi          = {10.48550/arXiv.1802.03426},
  note         = {arXiv preprint},
}

@article{Hoellt2019FC,
  title        = {{Focus+Context Exploration of Hierarchical Embeddings}},
  author       = {H\"{o}llt, T. and Vilanova, A. and Pezzotti, N. and Lelieveldt, B.P.F. and Hauser, H.},
  year         = 2019,
  journal      = {Computer Graphics Forum},
  volume       = 38,
  number       = 3,
  pages        = {569--579},
  doi          = {10.1111/cgf.13711},
  issn         = {1467-8659}
}

@article{Paulovich2008,
  title        = {{HiPP}: {A Novel Hierarchical Point Placement Strategy and its Application to the Exploration of Document Collections}},
  author       = {F.V. Paulovich and R. Minghim},
  year         = 2008,
  journal      = {{IEEE} Transactions on Visualization and Computer Graphics},
  volume       = 14,
  number       = 6,
  pages        = {1229--1236},
  doi          = {10.1109/tvcg.2008.138}
}

@article{ingramGlimmerMultilevelMDS2009,
  title        = {Glimmer: {{Multilevel MDS}} on the {{GPU}}},
  shorttitle   = {Glimmer},
  author       = {Ingram, Stephen and Munzner, Tamara and Olano, Marc},
  year         = 2009,
  journal      = {{IEEE} Transactions on Visualization and Computer Graphics},
  volume       = 15,
  number       = 2,
  pages        = {249--261},
  doi          = {10.1109/TVCG.2008.85}
}

@article{Marcilio-Jr2021HUMAP,
  title        = {{{HUMAP}}: {{Hierarchical Uniform Manifold Approximation}} and {{Projection}}},
  author       = {{Marc{\'i}lio-Jr}, Wilson E. and Eler, Danilo M. and Paulovich, Fernando V. and Martins, Rafael M.},
  year         = 2025,
  journal      = {IEEE Transactions on Visualization and Computer Graphics},
  volume       = 31,
  number       = 9,
  pages        = {5741-–5753},
  doi          = {10.1109/tvcg.2024.3471181}
}

@article{Pezzotti16HSNE,
  title        = {{Hierarchical Stochastic Neighbor Embedding}},
  author       = {Pezzotti, N. and H\"{o}llt, T. and Lelieveldt, B. and Eisemann, E. and Vilanova, A.},
  year         = 2016,
  journal      = {Computer Graphics Forum},
  volume       = 35,
  number       = 3,
  pages        = {21--30},
  doi          = {10.1111/cgf.12878}
}

@inproceedings{Vieth23,
  title        = {{Interactions for Seamlessly Coupled Exploration of High-Dimensional Images and Hierarchical Embeddings}},
  author       = {Vieth, Alexander and Lelieveldt, Boudewijn and Eisemann, Elmar and Vilanova, Anna and H\"{o}llt, Thomas},
  year         = 2023,
  booktitle    = {Proceedings of Vision, Modeling, and Visualization},
  publisher    = {The Eurographics Association},
  doi          = {10.2312/vmv.20231227},
}

@article{Nonato2019,
  title        = {Multidimensional {{Projection}} for {{Visual Analytics}}: {{Linking Techniques}} with {{Distortions}}, {{Tasks}}, and {{Layout Enrichment}}},
  author       = {Nonato, Luis Gustavo and Aupetit, Micha\"{e}l},
  year         = 2019,
  journal      = {IEEE Transactions on Visualization and Computer Graphics},
  volume       = 25,
  number       = 8,
  pages        = {2650--2673},
  doi          = {10.1109/TVCG.2018.2846735}
}

@article{Espadoto2021,
  title        = {Toward a {{Quantitative Survey}} of {{Dimension Reduction Techniques}}},
  author       = {Espadoto, Mateus and Martins, Rafael M. and Kerren, Andreas and Hirata, Nina S. T. and Telea, Alexandru C.},
  year         = 2021,
  journal      = {IEEE Transactions on Visualization and Computer Graphics},
  volume       = 27,
  number       = 3,
  pages        = {2153--2173},
  doi          = {10.1109/TVCG.2019.2944182}
}

@article{Kobak2019,
  title        = {The Art of Using T-{{SNE}} for Single-Cell Transcriptomics},
  author       = {Kobak, Dmitry and Berens, Philipp},
  year         = 2019,
  journal      = {Nature Communications},
  publisher    = {Nature Publishing Group},
  volume       = 10,
  number       = 1,
  pages        = 5416,
  doi          = {10.1038/s41467-019-13056-x},
  issn         = {2041-1723}
}

@article{Malkov2020,
  title        = {Efficient and {{Robust Approximate Nearest Neighbor Search Using Hierarchical Navigable Small World Graphs}}},
  author       = {Malkov, Yu A. and Yashunin, D. A.},
  year         = 2020,
  journal      = {IEEE Transactions on Pattern Analysis and Machine Intelligence},
  publisher    = {IEEE Computer Society},
  volume       = 42,
  number       = 4,
  pages        = {824--836},
  doi          = {10.1109/TPAMI.2018.2889473},
  issn         = 19393539
}

@misc{Crane2020,
  title        = {A {{Survey}} of {{Algorithms}} for {{Geodesic Paths}} and {{Distances}}},
  author       = {Crane, Keenan and Livesu, Marco and Puppo, Enrico and Qin, Yipeng},
  year         = 2020,
  doi          = {10.48550/arXiv.2007.10430},
  note         = {arXiv preprint},
}

@article{Gobel1974,
  title        = {Random Walks on Graphs},
  author       = {G{\"o}bel, F. and Jagers, A. A.},
  year         = 1974,
  journal      = {Stochastic Processes and their Applications},
  volume       = 2,
  number       = 4,
  pages        = {311--336},
  doi          = {10.1016/0304-4149(74)90001-5},
  issn         = {0304-4149}
}

@article{khosla2021,
  title        = {A {{Comparative Study}} for {{Unsupervised Network Representation Learning}}},
  author       = {Khosla, Megha and Setty, Vinay and Anand, Avishek},
  year         = 2021,
  journal      = {IEEE Transactions on Knowledge and Data Engineering},
  volume       = 33,
  number       = 5,
  pages        = {1807--1818},
  doi          = {10.1109/TKDE.2019.2951398}
}

@article{kruigerGraphLayoutsTSNE2017,
  title        = {Graph {{Layouts}} by T-{{SNE}}},
  author       = {Kruiger, J. F. and Rauber, P. E. and Martins, R. M. and Kerren, A. and Kobourov, S. and Telea, A. C.},
  year         = 2017,
  journal      = {Computer Graphics Forum},
  volume       = 36,
  number       = 3,
  pages        = {283--294},
  doi          = {10.1111/cgf.13187}
}

@inproceedings{groverNode2vec2016,
  title        = {Node2vec: {{Scalable Feature Learning}} for {{Networks}}},
  shorttitle   = {Node2vec},
  author       = {Grover, Aditya and Leskovec, Jure},
  year         = 2016,
  booktitle    = {Proceedings of {KDD}},
  pages        = {855--864},
  doi          = {10.1145/2939672.2939754},
  isbn         = {978-1-4503-4232-2}
}

@article{Lafon2006,
  title        = {Diffusion Maps and Coarse-Graining: {{A}} Unified Framework for Dimensionality Reduction, Graph Partitioning, and Data Set Parameterization},
  author       = {Lafon, St{\'e}phane and Lee, Ann B.},
  year         = 2006,
  journal      = {IEEE Transactions on Pattern Analysis and Machine Intelligence},
  volume       = 28,
  number       = 9,
  pages        = {1393--1403},
  doi          = {10.1109/TPAMI.2006.184},
  issn         = {01628828}
}

@article{VanderMaaten2008,
  title        = {Visualizing {{Data}} Using T-{{SNE}}},
  author       = {{van der Maaten}, Laurens and Hinton, G.E.},
  year         = 2008,
  journal      = {Journal of Machine Learning Research},
  volume       = 9,
  number       = 86,
  pages        = {2579--2605},
  issn         = {1533-7928},
  url          = {https://jmlr.org/papers/v9/vandermaaten08a.html}
}

@article{leeNonlinearDimensionalityReduction2005,
  title        = {Nonlinear Dimensionality Reduction of Data Manifolds with Essential Loops},
  author       = {Lee, John Aldo and Verleysen, Michel},
  year         = 2005,
  journal      = {Neurocomputing},
  series       = {Geometrical {{Methods}} in {{Neural Networks}} and {{Learning}}},
  volume       = 67,
  pages        = {29--53},
  doi          = {10.1016/j.neucom.2004.11.042},
  issn         = {0925-2312}
}

@article{gradyRandomWalksImage2006,
  title        = {Random {{Walks}} for {{Image Segmentation}}},
  author       = {Grady, L.},
  year         = 2006,
  journal      = {IEEE Transactions on Pattern Analysis and Machine Intelligence},
  volume       = 28,
  number       = 11,
  pages        = {1768--1783},
  doi          = {10.1109/TPAMI.2006.233},
  issn         = {1939-3539}
}

@article{shenLazyRandomWalks2014,
  title        = {Lazy {{Random Walks}} for {{Superpixel Segmentation}}},
  author       = {Shen, Jianbing and Du, Yunfan and Wang, Wenguan and Li, Xuelong},
  year         = 2014,
  journal      = {IEEE Transactions on Image Processing},
  volume       = 23,
  number       = 4,
  pages        = {1451--1462},
  doi          = {10.1109/TIP.2014.2302892},
  issn         = {1941-0042}
}

@article{kangDynamicRandomWalk2020,
  title        = {Dynamic {{Random Walk}} for {{Superpixel Segmentation}}},
  author       = {Kang, Xuejing and Zhu, Lei and Ming, Anlong},
  year         = 2020,
  journal      = {IEEE Transactions on Image Processing},
  volume       = 29,
  pages        = {3871--3884},
  doi          = {10.1109/TIP.2020.2967583},
  issn         = {1941-0042}
}

@article{wangSuperpixelSegmentationBenchmark2017,
  title        = {Superpixel Segmentation: {{A}} Benchmark},
  shorttitle   = {Superpixel Segmentation},
  author       = {Wang, Murong and Liu, Xiabi and Gao, Yixuan and Ma, Xiao and Soomro, Nouman Q.},
  year         = 2017,
  journal      = {Signal Processing: Image Communication},
  volume       = 56,
  pages        = {28--39},
  doi          = {10.1016/j.image.2017.04.007},
  issn         = {0923-5965}
}

@article{barcelosComprehensiveReviewNew2024,
  title        = {A {{Comprehensive Review}} and {{New Taxonomy}} on {{Superpixel Segmentation}}},
  author       = {Barcelos, Isabela Borlido and Bel{\'e}m, Felipe De Castro and Jo{\~a}o, Leonardo De Melo and Patroc{\'i}nio, Zenilton K. G. Do and Falc{\~a}o, Alexandre Xavier and Guimar{\~a}es, Silvio Jamil Ferzoli},
  year         = 2024,
  journal      = {ACM Comput. Surv.},
  volume       = 56,
  number       = 8,
  pages        = {200:1--200:39},
  doi          = {10.1145/3652509},
  issn         = {0360-0300}
}

@article{Tenenbaum2000,
  title        = {A Global Geometric Framework for Nonlinear Dimensionality Reduction},
  author       = {Tenenbaum, J. B. and De Silva, V. and Langford, J. C.},
  year         = 2000,
  journal      = {Science},
  volume       = 290,
  number       = 5500,
  pages        = {2319--2323},
  doi          = {10.1126/science.290.5500.2319},
  issn         = {00368075}
}

@article{achantaSLIC2012,
  title        = {{{SLIC}} Superpixels Compared to State-of-the-Art Superpixel Methods},
  author       = {Achanta, Radhakrishna and Shaji, Appu and Smith, Kevin and Lucchi, Aurelien and Fua, Pascal and S\"{u}sstrunk, Sabine},
  year         = 2012,
  journal      = {IEEE Transactions on Pattern Analysis and Machine Intelligence},
  volume       = 34,
  number       = 11,
  pages        = {2274--2282},
  doi          = {10.1109/TPAMI.2012.120}
}

@article{stutzSuperpixelsEvaluationStateoftheart2018,
  title        = {Superpixels: {{An}} Evaluation of the State-of-the-Art},
  shorttitle   = {Superpixels},
  author       = {Stutz, David and Hermans, Alexander and Leibe, Bastian},
  year         = 2018,
  journal      = {Computer Vision and Image Understanding},
  volume       = 166,
  pages        = {1--27},
  doi          = {10.1016/j.cviu.2017.03.007}
}

@misc{IndianPinesData,
  title        = {{220 Band AVIRIS Hyperspectral Image Data Set}: {June 12, 1992 Indian Pine Test Site 3}},
  author       = {Baumgardner, Marion F. and Biehl, Larry L. and Landgrebe, David A.},
  year         = 2015,
  doi          = {10.4231/R7RX991C}
}

@inproceedings{gualtieri1999,
  title        = {{Support Vector Machines for Hyperspectral Remote Sensing Classification}},
  author       = {Gualtieri, J. Anthony and Cromp, Robert F.},
  year         = 1999,
  booktitle    = {Proceedings of {{Advances}} in {{Computer-Assisted Recognition}}},
  publisher    = {{SPIE}},
  volume       = 3584,
  pages        = {221--232},
  doi          = {10.1117/12.339824}
}

@article{cycif,
  title        = {Highly multiplexed immunofluorescence imaging of human tissues and tumors using t-CyCIF and conventional optical microscopes},
  author       = {Lin, Jia-Ren and Izar, Benjamin and Wang, Shu and Yapp, Clarence and Mei, Shaolin and Shah, Parin M and Santagata, Sandro and Sorger, Peter K},
  year         = 2018,
  journal      = {eLife},
  publisher    = {eLife Sciences Publications, Ltd},
  volume       = 7,
  pages        = {e31657},
  doi          = {10.7554/eLife.31657},
  issn         = {2050-084X}
}

@article{Yapp2023,
  title        = {Highly multiplexed 3D profiling of cell states and immune niches in human tumors},
  author       = {Yapp, Clarence and Nirmal, Ajit J. and Zhou, Felix and Wong, Alex Y. H. and Tefft, Juliann B. and Lu, Yi Daniel and Shang, Zhiguo and Maliga, Zoltan and Llopis, Paula Montero and Murphy, George F. and Lian, Christine G. and Danuser, Gaudenz and Santagata, Sandro and Sorger, Peter K.},
  year         = 2025,
  journal      = {Nature Methods},
  volume       = 22,
  number       = 10,
  pages        = {2180--2193},
  doi          = {10.1038/s41592-025-02824-x},
  issn         = {1548-7105}
}

@misc{douze2024faiss,
  title        = {The Faiss library},
  author       = {Matthijs Douze and Alexandr Guzhva and Chengqi Deng and Jeff Johnson and Gergely Szilvasy and Pierre-Emmanuel Mazaré and Maria Lomeli and Lucas Hosseini and Hervé Jégou},
  year         = 2024,
  doi          = {10.48550/arXiv.2401.08281},
  note         = {arXiv preprint},
}

@article{Moon2019,
  title        = {Visualizing Structure and Transitions in High-Dimensional Biological Data},
  author       = {Moon, Kevin R. and {van Dijk}, David and Wang, Zheng and Gigante, Scott and Burkhardt, Daniel B. and Chen, William S. and Yim, Kristina and van den Elzen, Antonia and Hirn, Matthew J. and Coifman, Ronald R. and Ivanova, Natalia B. and Wolf, Guy and Krishnaswamy, Smita},
  year         = 2019,
  journal      = {Nature Biotechnology},
  publisher    = {Nature Research},
  volume       = 37,
  number       = 12,
  pages        = {1482--1492},
  doi          = {10.1038/s41587-019-0336-3},
  issn         = 15461696
}

@article{kuchroo2022,
  title        = {Multiscale {{PHATE}} Identifies Multimodal Signatures of {{COVID-19}}},
  author       = {Kuchroo, Manik and Huang, Jessie and Wong, Patrick and Grenier, Jean-Christophe and Shung, Dennis and Tong, Alexander and Lucas, Carolina and Klein, Jon and Burkhardt, Daniel B. and Gigante, Scott and Godavarthi, Abhinav and Rieck, Bastian and Israelow, Benjamin and Simonov, Michael and Mao, Tianyang and Oh, Ji Eun and Silva, Julio and Takahashi, Takehiro and Odio, Camila D. and {Casanovas-Massana}, Arnau and Fournier, John and Farhadian, Shelli and Dela Cruz, Charles S. and Ko, Albert I. and Hirn, Matthew J. and Wilson, F. Perry and Hussin, Julie G. and Wolf, Guy and Iwasaki, Akiko and Krishnaswamy, Smita},
  year         = 2022,
  journal      = {Nature Biotechnology},
  publisher    = {Nature Publishing Group},
  volume       = 40,
  number       = 5,
  pages        = {681--691},
  doi          = {10.1038/s41587-021-01186-x},
  issn         = {1546-1696}
}

@inproceedings{Bernard2015,
  title        = {{A Survey and Task-Based Quality Assessment of Static {{2D}} Colormaps}},
  author       = {Bernard, J{\"u}rgen and Steiger, Martin and Mittelst{\"a}dt, Sebastian and Thum, Simon and Keim, Daniel and Kohlhammer, J{\"o}rn},
  year         = 2015,
  booktitle    = {Proceedings of Visualization and {{Data Analysis}}},
  publisher    = {{SPIE}},
  volume       = 9397,
  pages        = {93970M},
  doi          = {10.1117/12.2079841},
  editor       = {Kao, David L. and Hao, Ming C. and Livingston, Mark A. and Wischgoll, Thomas}
}

@article{vkt2023manivault,
  title        = {ManiVault: A Flexible and Extensible Visual Analytics Framework for High-Dimensional Data},
  author       = {Vieth, Alexander and Kroes, Thomas and Thijssen, Julian and van Lew, Baldur and Eggermont, Jeroen and Basu, Soumyadeep and Eisemann, Elmar and Vilanova, Anna and Höllt, Thomas and Lelieveldt, Boudewijn},
  year         = 2024,
  journal      = {IEEE Transactions on Visualization and Computer Graphics},
  volume       = 30,
  number       = 1,
  pages        = {175--185},
  doi          = {10.1109/TVCG.2023.3326582}
}

@article{barbato22,
  title        = {Unsupervised segmentation of hyperspectral remote sensing images with superpixels},
  author       = {Mirko Paolo Barbato and Paolo Napoletano and Flavio Piccoli and Raimondo Schettini},
  year         = 2022,
  journal      = {Remote Sensing Applications: Society and Environment},
  volume       = 28,
  pages        = 100823,
  doi          = {https://doi.org/10.1016/j.rsase.2022.100823},
  issn         = {2352-9385}
}

@inproceedings{perozziDeepWalk2014,
  title        = {{{DeepWalk}}: Online Learning of Social Representations},
  shorttitle   = {{DeepWalk}},
  author       = {Perozzi, Bryan and {Al-Rfou}, Rami and Skiena, Steven},
  year         = 2014,
  booktitle    = {Proceedings of{{KDD}}},
  publisher    = {Association for Computing Machinery},
  address      = {New York, NY, USA},
  pages        = {701--710},
  doi          = {10.1145/2623330.2623732},
  isbn         = {978-1-4503-2956-9}
}

@inproceedings{kim2025,
  title        = {Revisiting Random Walks for Learning on Graphs},
  author       = {Kim, Jinwoo and Zaghen, Olga and Suleymanzade, Ayhan and Ryou, Youngmin and Hong, Seunghoon},
  year         = 2025,
  booktitle    = {Proceedings of ICLR},
  pages        = {82497--82547},
  note         = {arXiv document: \href{https://doi.org/10.48550/arXiv.2407.01214}{2407.01214}}
}

@incollection{Lovasz1993,
  title        = {Random Walks on Graphs: A Survey},
  author       = {László Lovász},
  year         = 1993,
  booktitle    = {Combinatorics, Paul Erdős is eighty},
  publisher    = {János Bolyai Mathematical Society},
  volume       = 2,
  pages        = {1--46},
  editor       = {D. Miklós and V. T. Sós and T. Szőnyi}
}

@article{Pezzotti2017,
  title        = {Approximated and User Steerable {{tSNE}} for Progressive Visual Analytics},
  author       = {Pezzotti, Nicola and Lelieveldt, Boudewijn P.F. and Van Der Maaten, Laurens and H{\"o}llt, Thomas and Eisemann, Elmar and Vilanova, Anna},
  year         = 2017,
  journal      = {IEEE Transactions on Visualization and Computer Graphics},
  publisher    = {IEEE Computer Society},
  volume       = 23,
  number       = 7,
  pages        = {1739--1752},
  doi          = {10.1109/TVCG.2016.2570755},
  issn         = 10772626
}

@misc{BioMedVisChallenge2025,
  title        = {{Bio+MedVis Challenge at IEEE VIS 2025}},
  author       = {BioVis},
  year         = 2025,
  howpublished = {URL: \href{https://web.archive.org/web/20250929182600/https://biovis.net/2025/biovisChallenges\_vis/}{archived webpage}, \href{https://biovis.net/2025/biovisChallenges\_vis/}{\mbox{biovis}.net/2025/biovisChallenges\_vis}}
}

@misc{SalinasData,
  title        = {Hyperspectral Remote Sensing Scenes},
  author       = {M. Graña, MA. Veganzons, B. Ayerdi},
  howpublished = {URL: \href{https://web.archive.org/web/20250926223435/https://www.ehu.eus/ccwintco/index.php/Hyperspectral\_Remote\_Sensing\_Scenes/}{archived webpage}, \href{http://www.ehu.eus/ccwintco/index.php?title=Hyperspectral\_Remote\_Sensing\_Scenes\#Salinas-A\_scene}{\fontsize{6.85}{8}\selectfont ehu.eus/ccwintco/index.php?title=Hyperspectral\_Remote\_Sensing\_Scenes}},
  year         = 1998
}

@article{Roweis2000,
  title        = {Nonlinear dimensionality reduction by locally linear embedding},
  author       = {Roweis, S. T. and Saul, L. K.},
  year         = 2000,
  journal      = {Science},
  volume       = 290,
  number       = 5500,
  pages        = {2323--2326},
  doi          = {10.1126/science.290.5500.2323}
}

@article{Warchol2025SEAL,
  title        = {SEAL: Spatially-resolved Embedding Analysis with Linked Imaging Data},
  author       = {Warchol, Simon and Guo, Grace and Knittel, Johannes and Freeman, Dan and Bhalla, Usha and Muhlich, Jeremy L and Sorger, Peter K. and Pfister, Hanspeter},
  year         = {2026},
  volume       = {32},
  number       = {1},
  pages        = {484-494},
  journal      = {IEEE Transactions on Visualization and Computer Graphics}, 
  doi          = {10.1109/TVCG.2025.3634794}
}

@article{Pan2025Tregs,
  title = {Regulatory T cells in solid tumor immunotherapy: effect,  mechanism and clinical application},
  volume = {16},
  DOI = {10.1038/s41419-025-07544-w},
  number = {1},
  journal = {Cell Death \& Disease},
  publisher = {Springer Science and Business Media LLC},
  author = {Pan,  Yan and Zhou,  Hanqiong and Sun,  Zhenqiang and Zhu,  Yichen and Zhang,  Zhe and Han,  Jing and Liu,  Yang and Wang,  Qiming},
  year = {2025},
}

\appendix 
\begin{strip}
\centering
{ \bfseries
    { \Large
        Supplemental Material: \\ \vspace{8pt}
    }
    { \huge
        \mainPaperTitle  \vspace{14pt}
    }
}

\end{strip}

\section{Symmetrized and connected $k$NN-graph} \label{supp:symKnn}
We convert the directed $k$NN-graph into an undirected graph by making each edge bidirectional, compute its connected components and add new edges that connect the components. %
For that purpose, we first compute the mean value of the high-dimensional data points in each connected component. 
Secondly, we set up a fully-connected helper graph, each vertex corresponding to one connected component and edge weights being the $\delta$~distances between their means.
We then compute its minimum spanning tree (MST).
Finally, we find the smallest $\delta$~distance between any two points from two components connected by an edge in the above MST and introduces a bidirectional connection into the symmetrized $k$NN-graph.
If a symmetrized $k$NN-graph already only has one connected component, no new edges are introduced.

\section{Geodesic distance approach} \label{supp:geodesic}
Their geodesic distance~$d_{\graph}(\ID, \ID[j])$ between two data points is their geodesic distance, i.e., the accumulated distance along the shortest-paths between their vertices in~\graph.
Since on the data level every component~$\component_p^0 \in \superpixel^{0}$ is a single pixel and corresponds directly to a single data point~\ID\ we can immediately apply the geodesic distance to compare them. 
On higher levels though, we need to broaden the definition of this measure since there we are associating sets of pixels/data points.
One successfully applied measure %
used for comparing sets of points is the Hausdorff distance:
\begin{equation}
	d_H(\component_p, \component_q) = \max \biggl\{ 
	\max_{\ID \in \component_p} \left( \min_{\ID[j] \in \component_q} d(\ID, \ID[j]) \right),\ 
	\max_{\ID[j] \in \component_q} \left( \min_{\ID \in \component_p} d(\ID[j], \ID) \right)
	\biggr\}
	\label{eq:haus}
\end{equation}
with $d(\ID, \ID[j]) = d_{\graph}(\ID, \ID[j])$.
That is, we capture the worst-case mismatch between the sets by finding the maximum distance from any pixel in one superpixel to any pixel in the other. 
When both superpixel just contain a single pixel, $d_H$ simplifies to the geodesic distance $d_{\graph}$ between them.

Both the construction of a search index in HNSW and the actual nearest neighbor search is rather feasible with complexities of respectively $O(n \log n)$ and $O(\log n)$. 
But computing the geodesic distance~$d_{\graph}$ between two graph nodes is intrinsically complex and increases the typical complexity of distance computation for, e.g., Euclidean distances, of $O(1)$ to 
$O((n + |\edges|) \log n )$ 
with the A* search algorithm or
$O(n \log n + |\edges|)$ 
with Dijkstra's algorithm.
This distance needs to be computed~$|\component_p| \cdot |\component_q|$ times for the Hausdorff-based set comparison.
On higher levels the components contain numbers of pixels in the same order of magnitude as the entire image, such that computing a single set comparison with the Hausdorff distance will become intractable, as well.
One way to address this combinatorial burden emerging for large superpixels on higher abstraction levels could be taking a number of, e.g., 100, sample pixels from each superpixel and compute the Hausdorff distance between these representative subsets.
Figure~\ref{sup:fig:bus_geo} shows that this geodesics-based approaches can still yield decent results, here a small shown for a color image.
The embeddings therein are also not randomly initialized but instead based on the previous level's embedding using the average position of all corresponding merged points from the lower-level
embedding.

\section{Dissimilarity matrix derivation} \label{supp:derivations}

The formulation fo the dissimilarity matrix in \autoref{eq:probDistOnLevel} rewrites the summation over matrix entries from \autoref{eq:bc0} as a matrix multiplication.
This follows straight from definition of the matrix product.
Given that A is a $m \times n$ matrix, $C = A B$ with $B = A^\top$ and $\top$ indicating a transposition such that $\left[A^\top \right]_{ij} = \left[A \right]_{ji}$, the elements in the $m \times m$ matrix C are
\begin{align*}
    c_{ij} &= \sum_{k=1}^n a_{ik} b_{kj} = \sum_{k=1}^n a_{ik} a_{jk} \\
    &= \sum_{k=1}^n \sqrt{\transitions(\ID, \ID[k])} \sqrt{\transitions(\ID[j], \ID[k])} \\
    &= \sum_{k=1}^n \sqrt{\transitions(\ID, \ID[k])\ \transitions(\ID[j], \ID[k])}
\end{align*}
with $a_{ik} = \sqrt{\transitions(\ID, \ID[k])}$ using the same element-wise square root operation as in \autoref{sec:method:ComputeEmbeddings}.

\section{Implementation} \label{supp:implementation}

As the computation of the~$k$NN-graph on the data level tends to be increasingly expensive for large data sets, we instead approximate the k-nearest neighbors with HNSW~\cite{Malkov2020} using Faiss~\cite{douze2024faiss}.
Using an approximated~$k$NN-graph is common practice in virtually all contemporary neighborhood-based DR methods since its first introduction in A-tSNE~\cite{Pezzotti2017} as is comes with small to negligible loss of embedding quality.

The random walks on the data graph with~\randomWalkSteps\ steps are a Markov Chain Monte Carlo technique to approximate~$P^{\randomWalkSteps}$.
While it is possible to perform the sparse matrix multiplication explicitly, we chose to approximate it with the random walks.
This allows us to tweak the number of repeated walks~\randomWalkNumber\ and step length~\randomWalkSteps\ to adjust the tradeoff between computation speed, and approximation accuracy and memory consumption.

An implementation of our method as a standalone library and an interactive tool with coordinated views between image and embedding representation in the ManiVault framework~\cite{vkt2023manivault} are available at 
\href{https://github.com/anonymous/}{github.com/anonymous}%
.

\section{Indian Pines} \label{supp:pines_large}

Settings for the superpixel hierarchy discussed in Section~\ref*{sec:eval:hyper}: 
$k$NN algorithm: HNSW (\href{https://github.com/facebookresearch/faiss}{faiss} implementation), 
300 nearest neighbors, 
Number of random walks: 50, 
Length of random walks: 25.

Settings for the HSNE hierarchy (default settings of the implementation in the \href{https://github.com/biovault/HDILib/}{HDILib}:
$k$NN algorithm: HNSW (\href{https://github.com/nmslib/hnswlib}{hnswlib} implementation), 
perplexity of 30 (yielding 90 nearest neighbors),
Number of random walks (landmark selection): 15, 
Number of random walks (area of interest): 100, 
Length of random walks: 15.

Figure~\ref{fig:IndianPinesLargeRecolored} shows the superpixel hierarchy from Section~\ref*{sec:eval:hyper} for all level from 0 to 7.
It is possible to continue with more abstraction levels until there is only one superpixel, which covers the entire image, but the very large superpixles on high levels do not create interesting subdivision of the image anymore and are not shown here.

\section{CiCYF} \label{supp:cicyf}

The marker channels used for the superpixel hierarchy and corresponding embeddings for the CiCYF discussed in Section~\ref*{sec:eval:cycif} (as per personal correspondence with the data paper's authors):
5hmc, 
b-actin, 
CD103, 
CD11b, 
CD11c, 
CD163, 
CD20, 
CD206, 
CD31, 
CD3E, 
CD4, 
CD8a, 
FOXP3, 
HLA-AB, 
IRF1, 
Ki67, 
laminABC, 
MART1, 
panCK, 
PD1, 
podoplanin, 
S100A, 
S100B, 
SOX10, 
SOX9, 
pMLC2, 
yH2AX.

\begin{figure}[t]
    \centering
    \includegraphics[width=1\linewidth]{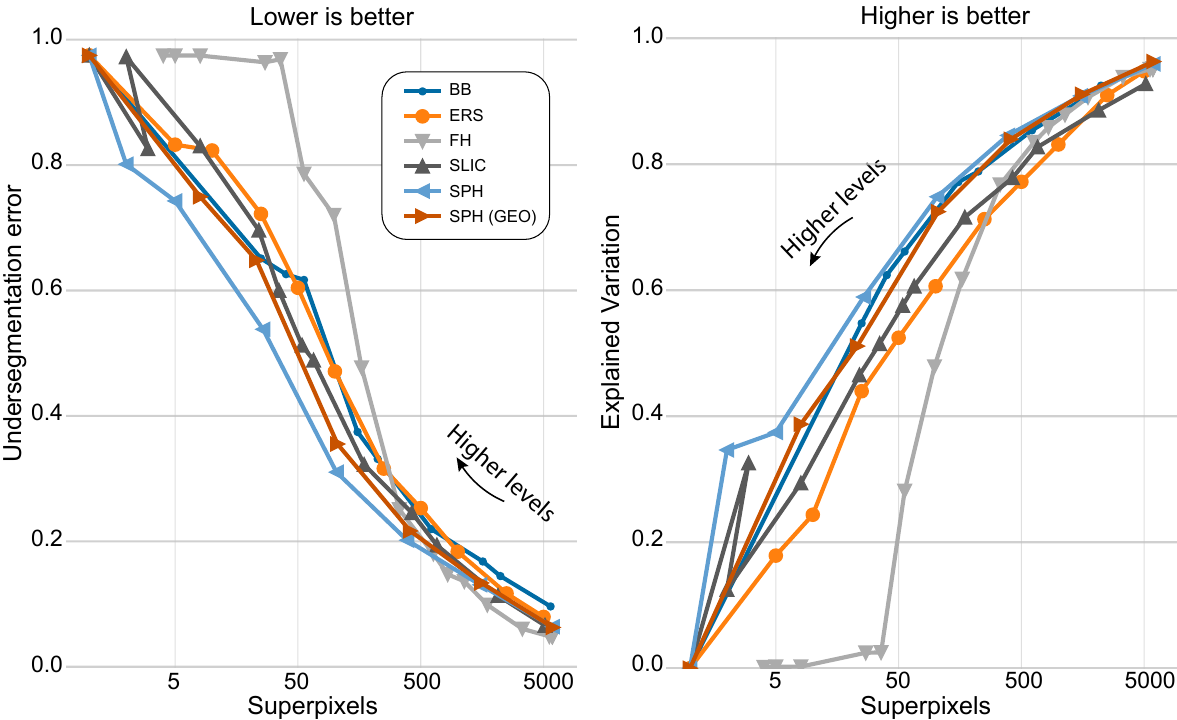}
    \caption{\textbf{Explained Variation and Undersegmentation Error:} %
    For Pines data.
    }%
    \label{fig:qaunt:results_pines}%
\end{figure}

The channels Collagen and Hoechst (a marker used to stain DNA and labeled as "DNA" in our figure) used for recoloring in Figure~\ref*{fig:cicyf:overview} are not part of the input markers to the superpixel hierarchy.

Figure~\ref{fig:cicyf:largeImage} shows the full image data and indicates the region we focus on in Section~\ref*{sec:eval:cycif}, with $x_{begin} = 2060$,  $x_{size} = 2000$, $y_{begin} = 450$, $y_{size} = 1500$.

\section{Quantitative results: Indian Pines dataset} \label{supp:quant}

We computed Undersegmentation Error (UE) and Explained variation (EV) for our analysis. 
UE measures how much a superpixel extends outside an overlapping ground truth segment. 
If all superpixels completely lie within ground truth segments, the UE is zero.
Consequently, larger superpixels on higher abstraction levels will cause an increasing UE.
Given a ground-truth segmentation~\superpixelGroundTruth\ and a superpixel segmentation~\superpixel, we use:
\begin{equation}
    UE(\superpixelGroundTruth, \superpixel) = 
    \frac{1}{|\superpixel|} 
    \sum_{\componentGroundTruth_i \in \superpixelGroundTruth} 
    \sum_{\substack{\component_r \in \superpixel\\ \component_r \cap\ \componentGroundTruth_i \neq \emptyset}} 
    \hspace{-0.225cm} 
    \min\{
    | \component_r \cap \componentGroundTruth_i |,      %
    | \component_r \setminus \componentGroundTruth_i |  %
    \},
    \label{eq:undersegerror}
\end{equation}
As before, level indices are omitted for readability, i.e., \superpixel\ may be from any abstraction level $l$.
Lower UE is considered better

\begin{figure}[t]
    \centering
    \includegraphics[width=1\linewidth]{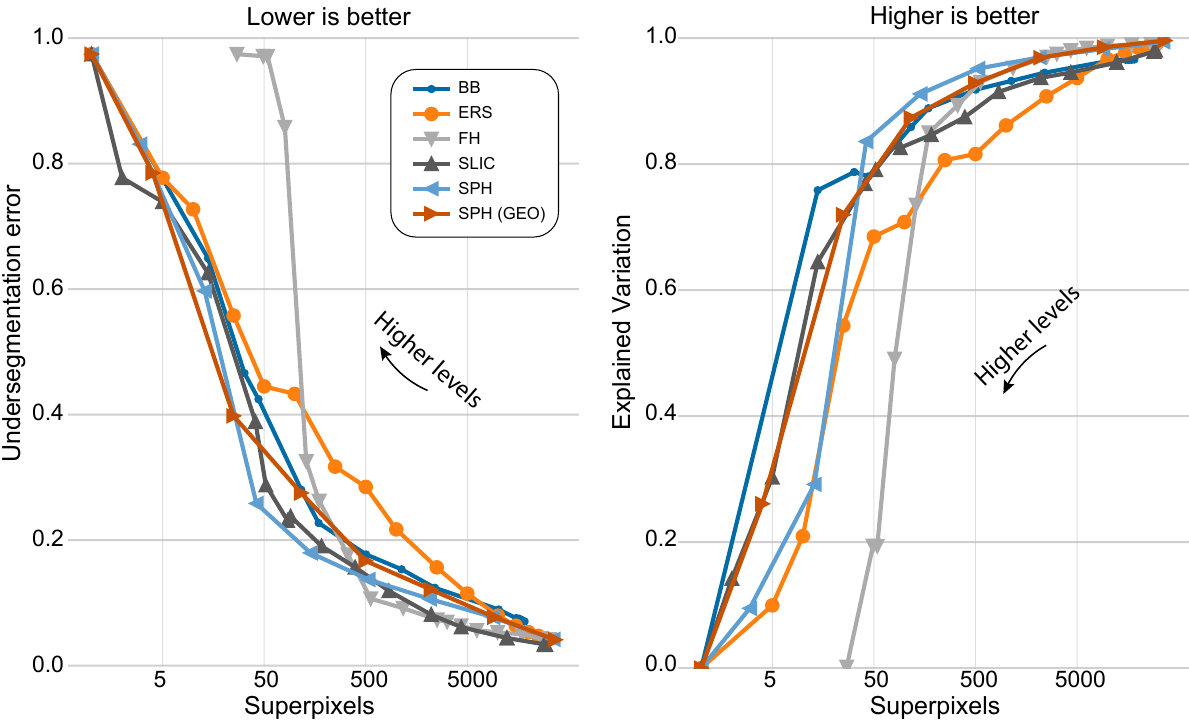}
    \caption{\textbf{Explained Variation and Undersegmentation Error:} %
    For Salinas data.
    }%
    \label{fig:qaunt:results_salinas}%
\end{figure}

\begin{table}[t]
\centering
\caption{\textbf{Explained Variation and Undersegmentation Error:} normalized area under the log\textsubscript{10}-plotted UE and (1 - EV) value curves in $\%$, i.e. the area under the lines in \Autoref{fig:qaunt:results_pines, fig:qaunt:results_salinas}. Lower values are better for both. Note, that the FH curves do not start at 1, which skews these aggregate measure in its favor.}
\label{tab:quant_auc_log}
\begin{tabular}{lS[table-format=2.2]S[table-format=2.2]S[table-format=2.2]S[table-format=2.2]}
\toprule
& \multicolumn{2}{c}{Pines {\small\cite{IndianPinesData}}} & \multicolumn{2}{c}{Salinas {\small\cite{SalinasData}}} \\
\midrule
\textbf{Method} & {log\textbf{AUE}} & {log\textbf{AEV}} & {log\textbf{AUE}} & {log\textbf{AEV}} \\
\midrule
FH {\small\cite{Felzenszwalb2004}} & {53.04} & {50.48}  & \textbf{26.85} & \textbf{19.28} \\ %
ERS {\small\cite{Liu2011Entropy}} & {52.64} & {47.80} & {41.66} & {37.30} \\ %
SLIC {\small\cite{achantaSLIC2012}} & {50.09} & {43.16} & {32.95} & {27.75} \\ %
BB {\small\cite{barbato22}} & {50.71} & {39.32} & {39.14} & {24.62} \\ %
SPH geodesic {\small\textit{(ours)}} & {46.56} & {38.94} & {34.05} & {25.86} \\
SPH $k$NN {\small\textbf{(ours)}} & \textbf{42.15} & \textbf{34.49} & {32.45} & {29.53} \\
\bottomrule
\end{tabular}
\end{table}

EV measures the superpixel quality without reference to a ground truth.
It tries to capture how much of the original image's pixel variation is preserved by the superpixel representation.
We define the mean attribute value for a channel~$c$ of a superpixel~$\component_r$, denoted~$\mean_c(\component_r)$, as the average of the channel attributes of all image pixels contained in the superpixel.
The global channel mean is referred to as~$\mean_c^{\imgGraph}$. 
As such, we use:
\begin{equation}
    EV(\superpixel) = 
    \frac
    {\sum_{c \in [1, \numAttrDimensions]} \sum_{\component_r \in \superpixel} |\component_r| (\mean_c(\component_r) - \mean_c^{\imgGraph})^2}
    {\sum_{c \in [1, \numAttrDimensions} \sum_{i \in [1, \imgSizeTot]} (\dataPointScalar_{i c} - \mean_c^{\imgGraph})^2}
    \label{eq:explainedvar}
\end{equation}
where~$\dataPointScalar_{i c}$ is the attribute value in image channel~$c$ for data point~$i$.
Higher EV is considered better.
In the main paper we consider the area under the curve given by (1 - EV) for multiple EV values to align with the UE metric.

Our method does not have a direct input parameter that steers the number of output superpixels.
Nonetheless, both our main method and its two variations yield a segmentation of around 5000 superpixels at the first abstraction level. 
Barbato's method steers the number of computed superpixels using an input parameter, \texttt{n\_clusters}.
We evaluate Barbato's method a range of \texttt{n\_clusters} that yields superpixel segmentations with similar numbers of superpixels as our abstraction levels. 
We use ${\texttt{m\_clust} = 0.8}$ as proposed by the authors as the suggested optimal setting.
Additionally, we set the input parameter \texttt{m} to 0, as any non-zero value enforced box-shaped superpixels and resulted in worse metrics.

Even though UE is constrained to $[0, 1]$ and a value of 1 can be expected for a worst-case segmentation, a "segmentation" of the Indian Pines data into a single segment yields a UE of $0.975$.
This is due to the term~$%
    \min\{%
    | \component_r \cap \componentGroundTruth_i |,      %
    | \component_r \setminus \componentGroundTruth_i |  %
    \}%
$
in equation~\ref*{eq:undersegerror}, which was introduced in this measure to improve in earlier formulations of undersegmentation error which tend to penalize small overlap of large superpixels with a ground truth segment.
However, since the ground-truth label "background" covers just-so more than half the image pixels, this is the only term in which $| \component_r \setminus \componentGroundTruth_i |$ is considered instead of $| \component_r \cap \componentGroundTruth_i |$.
This edge case of a ground-truth segment covering more than $50\%$ does not impact the evaluation though.

\paragraph*{Algorithm implementations}
We used the following implementations for the evaluation:
\begin{itemize}[nolistsep]
    \item FH: \href{https://scikit-image.org/docs/0.25.x/api/skimage.segmentation.html#skimage.segmentation.felzenszwalb}{scikit-image.org} (v0.25.2)
    \item ERS: \href{https://github.com/shubhranil04/S3PCA/blob/4a5e2c2bf69cc4c7b81187873ff2755476f12313/S3PCA_Experiments.ipynb}{github.com/shubhranil04/S3PCA} \\ (re-implementation of the reference implementation at \href{https://github.com/mingyuliutw/EntropyRateSuperpixel}{github.com/mingyuliutw/EntropyRateSuperpixel} allowing for hyperspectral images as input)
    \item SLIC: \href{https://scikit-image.org/docs/0.25.x/api/skimage.segmentation.html#skimage.segmentation.slic}{scikit-image.org} (v0.25.2)
    \item BB \href{https://github.com/alxvth/BarbatoHyperspectralSLIC/}{github.com/alxvth/BarbatoHyperspectralSLIC} \\ (fork of the reference implementation at \href{https://github.com/mpBarbato/Unsupervised-Segmentation-of-Hyperspectral-Remote-Sensing-Images-with-Superpixels}{github.com/mpBarbato/Unsupervised-Segmentation-of-Hyperspectral-Remote-Sensing-Images-with-Superpixels})
\end{itemize}

\begin{figure}[t]
    \centering
    \includegraphics[width=1\linewidth]{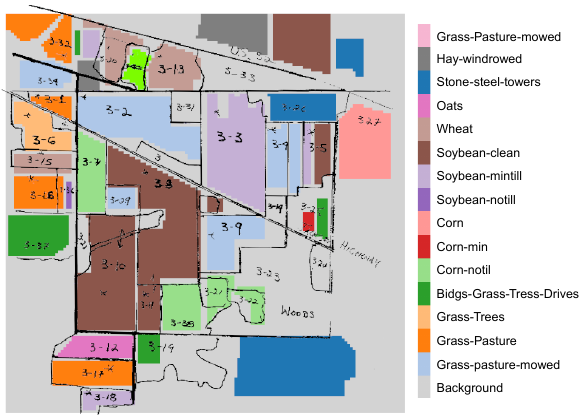}
    \caption{\textbf{Indian Pines Ground Truth:} %
    Ground Truth segmentation and labels with map overlay.
    }%
    \label{fig:pines:gt}%
\end{figure}

\paragraph*{Settings grid search}
We performed a grid search over relevant setting combinations.
We considered these settings:
\begin{itemize}[nolistsep]
    \item FH: $\sigma$ and $\texttt{min\_size}$, for preprocessing smoothing and postprocessing component size enforcement
    \item ERS: $\lambda$, weight of the balancing term (called \texttt{alpha} in the implementation)
    \item SLIC: $\sigma$ and $m$ (called $\texttt{compactness}$ in the implementation), for preprocessing smoothing and weighting spatial distance information
    \item BB: $m$ and $m_{\textrm{clust}}$, weighting spatial and clustered hyperspectral information respectively
    \item SPH: \randomWalkSteps, \randomWalkNumber\ and $\texttt{connectivity}$, the step lengths and number of random walks, and image-space connectivity (four or eight)
    \item SPH geodesic: $\texttt{connectivity}$
\end{itemize}

\begin{figure}[t]
    \centering
    \includegraphics[width=0.75\linewidth]{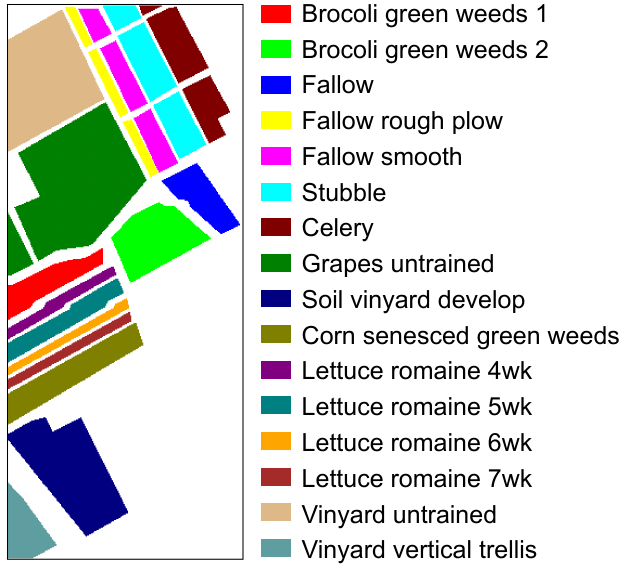}
    \caption{\textbf{Salinas Ground Truth:} %
    Ground Truth segmentation. White is background.
    }%
    \label{fig:salinas:gt}%
\end{figure}

\paragraph*{Algorithm settings}
We used the following settings for the evaluation, which yielded the lowest AUE. 
Listed in brackets is the parameter used for varying the superpixel size:
\begin{itemize}[nolistsep]
    \item FH: [$k$ (called $\texttt{scales}$ in the implementation)]
\begin{itemize}[nolistsep]
    \item Pines: $\sigma=1$, $\texttt{min\_size}=2$
    \item Salinas: $\sigma=1$, $\texttt{min\_size}=2$
\end{itemize}
    \item ERS: [$K$]
\begin{itemize}[nolistsep]
    \item Pines: $\lambda=0.2$
    \item Salinas: $\lambda=0.3$
\end{itemize}
    \item SLIC: [$k$ (called $\texttt{n\_segments}$ in the implementation)]
\begin{itemize}[nolistsep]
    \item Pines: $\sigma=1$ and $m=0.2$ 
    \item Salinas: $\sigma=3$ and $m=0.2$ 
\end{itemize}
    \item BB: [$K$ (called $\texttt{n\_cluster}$ in the implementation)]
\begin{itemize}[nolistsep]
    \item Pines: $m=0$ and $m_{\textrm{clust}}=0.6$
    \item Salinas: $m=0$ and $m_{\textrm{clust}}=5.0$
\end{itemize}
    \item SPH: [hierarchy levels]
\begin{itemize}[nolistsep]
    \item Pines: $\randomWalkSteps=10$, $\randomWalkNumber=50$ and $\texttt{connectivity}=4$
    \item Salinas: $\randomWalkSteps=10$, $\randomWalkNumber=100$ and $\texttt{connectivity}=8$
\end{itemize}
    \item SPH geodesic: [hierarchy levels]
\begin{itemize}[nolistsep]
    \item Pines: $\texttt{connectivity}=4$
    \item Salinas: $\texttt{connectivity}=8$
\end{itemize}
\end{itemize}
For unspecified settings we used the default values of the respective implementation.
Both SPH methods were computed with 90 nearest neighbors in the $k$NN graph.
SPH geodesic uses 100 samples as described above, see appendix~\ref{supp:geodesic}.
Note that BB performed slightly (but not significantly) better for $m_{\textrm{clust}}=0$ on the pines data, but we report results for the above non-zero $m_{\textrm{clust}}$ setting since otherwise an important part of the method would be lost. 

\begin{figure*}[t]
    \centering
    \includegraphics[width=1\linewidth]{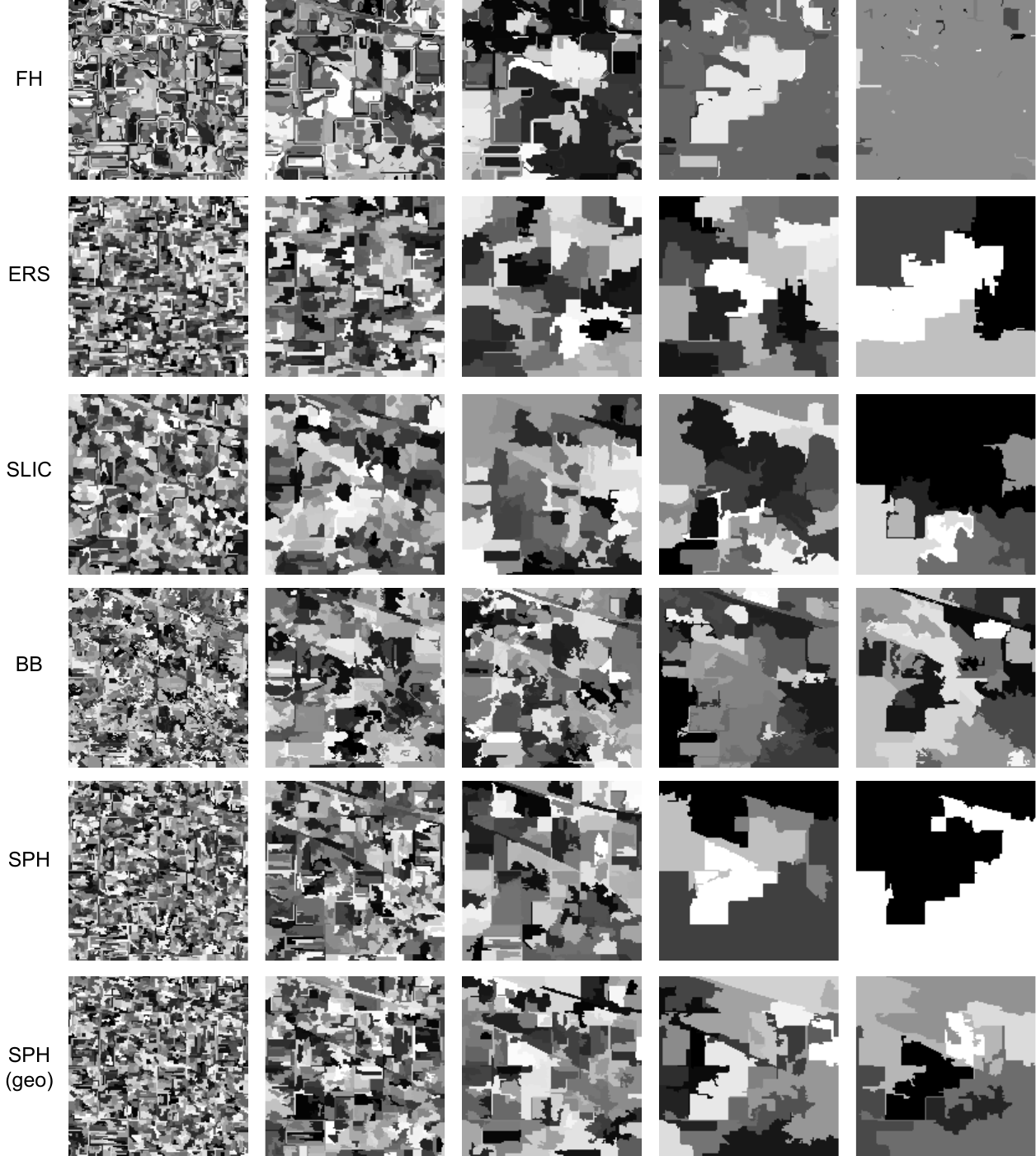}
    \caption{\textbf{Indian Pines comparison:} %
    Segmentations on a selection of representative levels for all evaluated methods. Each superpixel is randomly assigned a gray-value for distinction.
    }%
    \label{fig:comparison:pines}%
\end{figure*}

\begin{figure*}[t]
    \centering
    \includegraphics[width=1\linewidth]{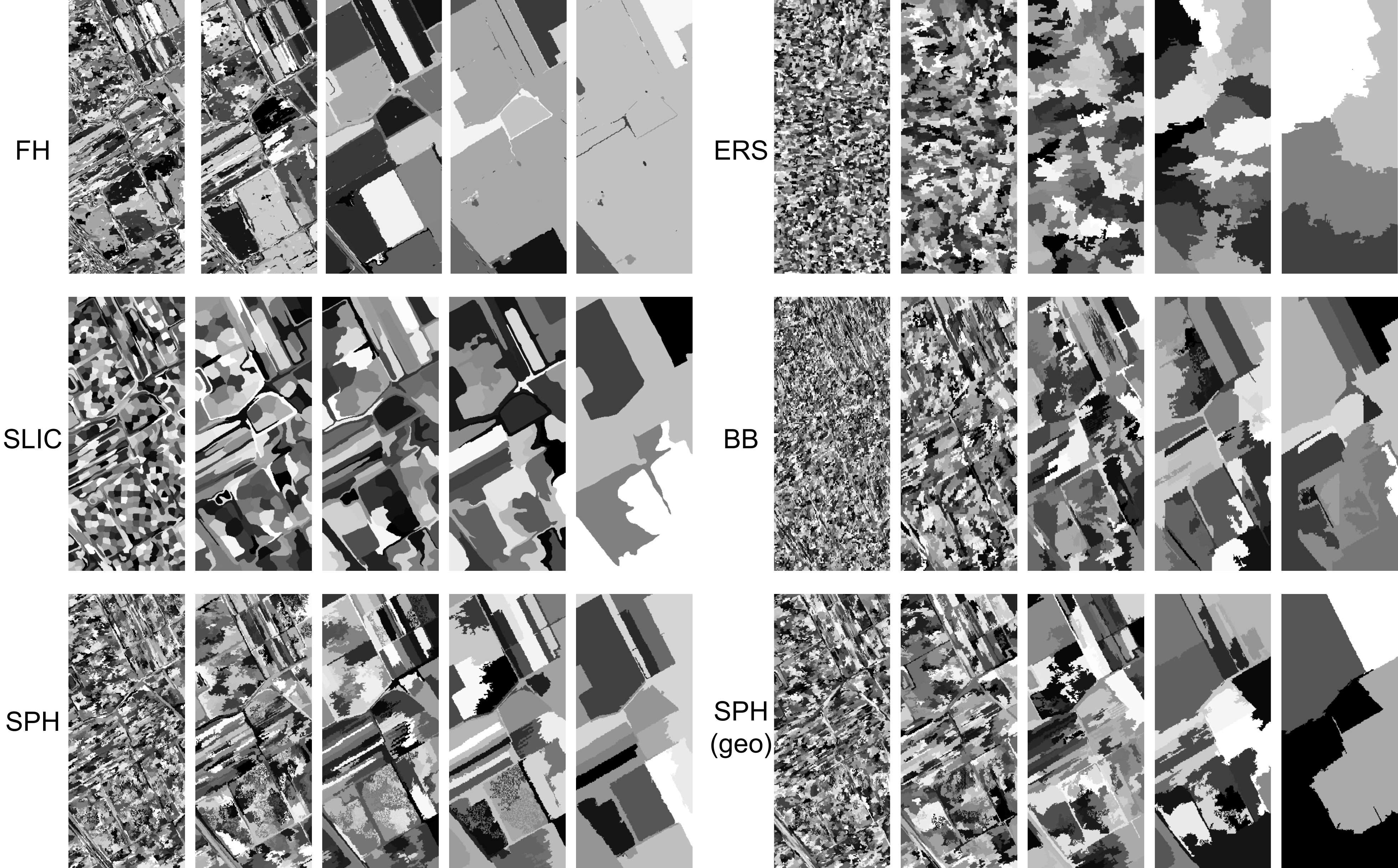}
    \caption{\textbf{Salinas comparison:} %
    Segmentations on a selection of representative levels for all evaluated methods. Each superpixel is randomly assigned a gray-value for distinction.
    }%
    \label{fig:comparison:salinas}%
\end{figure*}

Both SLIC and FH smoothing the input image as a preprocessing step (controlled by their $\sigma$ parameter).
An equivalent step could be performed for the other techniques as well, but in this evaluation we chose to use the available implementations without unnecessary modifications.

\begin{figure*}[t]
    \centering
    \includegraphics[width=\linewidth]{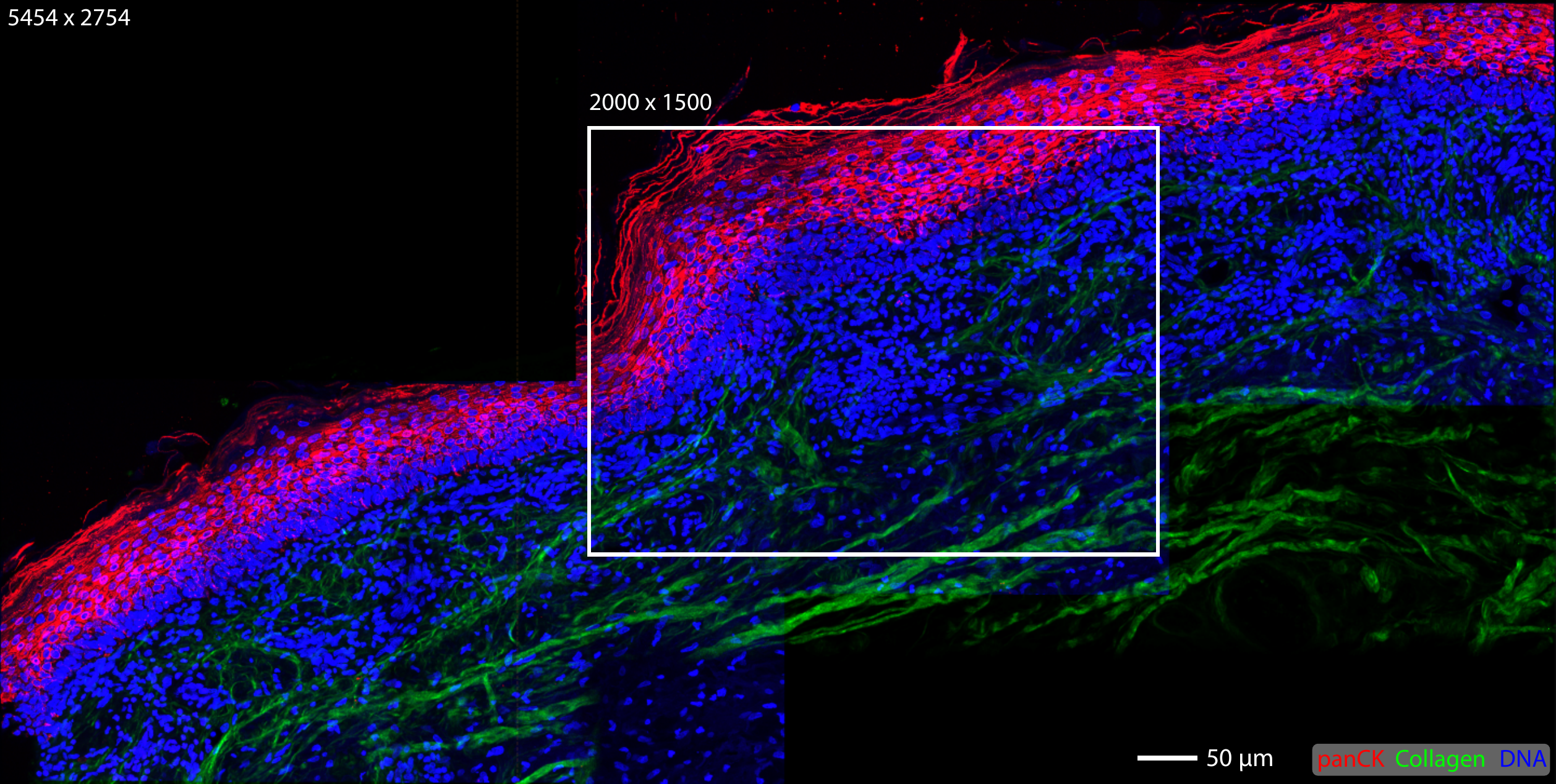}
    \vspace{-3mm}
    \caption{\textbf{CyCIF Focus Region:} %
    As used in section 6.2. 
    DNA corresponds to the Hoechst channel.
    }%
    \label{fig:cicyf:largeImage}%
    \vspace{0mm}
\end{figure*}

\begin{figure*}[t]
    \centering
    \includegraphics[width=0.74\linewidth]{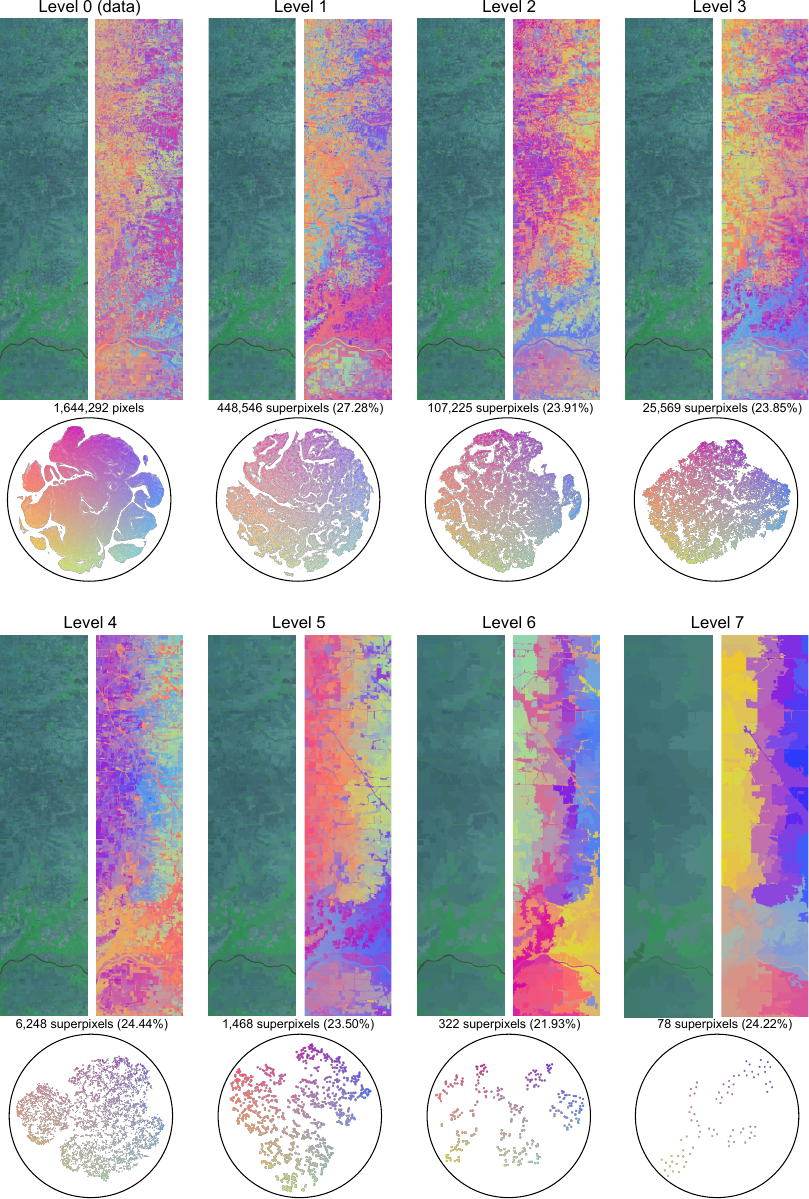}
    \caption{\textbf{Indian Pines superpixel hierarchy:} %
    Data level embedding alongside seven superpixel abstraction level embedding. 
    Each level shows a false color image based on the average spectrum per superpixel of channel 20 (587 nm, red), 76 (1090 nm, green) and 130 (1591 nm, blue). 
    Next to them are recolorings of the superpixel based on the embedding layout using a 2D colormap which is superimposed on the respective embeddings (as shown in \autoref{fig:IndianPinesLargeExploration}.
    Superpixels are hardly visible in these down-scaled version of the originally ${614 \times 2,678}$ (w x h) images, but more abstract levels clearly show more and more high-level structure.
    The percentages indicates the reduction of components, i.e., the number of components (superpixels) in level 1 reduces to 27.28\% of the previous level.
    In contrast to \autoref{fig:IndianPinesSmall}, embeddings starting from level 1 are initialized randomly.
    }%
    \label{fig:IndianPinesLargeRecolored}%
\end{figure*}

\begin{figure*}[t]
  \begin{subfigure}{0.14\textwidth}
    \centering
    \includegraphics[width=0.95\linewidth]{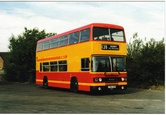}
    \label{sup:fig:bus_geo:img:0}
  \end{subfigure}%
  \hfill
  \begin{subfigure}{0.14\textwidth}
    \centering
    \includegraphics[width=0.95\linewidth]{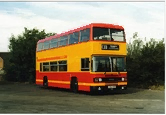}
    \label{sup:fig:bus_geo:img:1}
  \end{subfigure}%
  \hfill
  \begin{subfigure}{0.14\textwidth}
    \centering
    \includegraphics[width=0.95\linewidth]{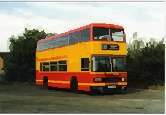}
    \label{sup:fig:bus_geo:img:2}
  \end{subfigure}%
  \hfill
  \begin{subfigure}{0.14\textwidth}
    \centering
    \includegraphics[width=0.95\linewidth]{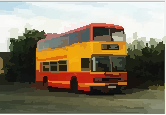}
    \label{sup:fig:bus_geo:img:3}
  \end{subfigure}%
  \hfill
  \begin{subfigure}{0.14\textwidth}
    \centering
    \includegraphics[width=0.95\linewidth]{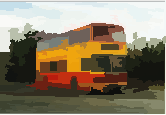}
    \label{sup:fig:bus_geo:img:4}
  \end{subfigure}%
  \hfill
  \begin{subfigure}{0.14\textwidth}
    \centering
    \includegraphics[width=0.95\linewidth]{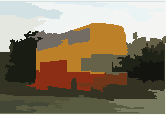}
    \label{sup:fig:bus_geo:img:5}
  \end{subfigure}%
  \hfill
  \begin{subfigure}{0.14\textwidth}
    \centering
    \includegraphics[width=0.95\linewidth]{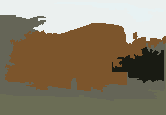}
    \label{sup:fig:bus_geo:img:6}
  \end{subfigure}
  \vspace{0mm}
  \begin{subfigure}{0.14\textwidth}
    \centering
    \includegraphics[width=.95\linewidth]{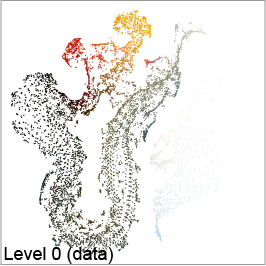}
    \label{sup:fig:bus_geo:emb:0}
  \end{subfigure}%
  \hfill
  \begin{subfigure}{0.14\textwidth}
    \centering
    \includegraphics[width=.95\linewidth]{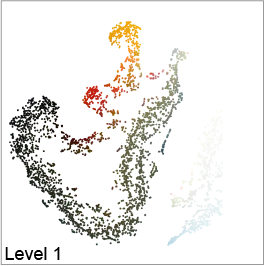}
    \label{sup:fig:bus_geo:emb:1}
  \end{subfigure}%
  \hfill
  \begin{subfigure}{0.14\textwidth}
    \centering
    \includegraphics[width=.95\linewidth]{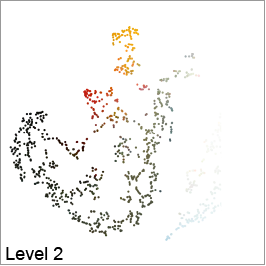}
    \label{sup:fig:bus_geo:emb:2}
  \end{subfigure}%
  \hfill
  \begin{subfigure}{0.14\textwidth}
    \centering
    \includegraphics[width=.95\linewidth]{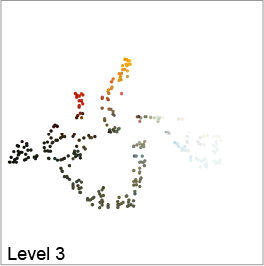}
    \label{sup:fig:bus_geo:emb:3}
  \end{subfigure}%
  \hfill
  \begin{subfigure}{0.14\textwidth}
    \centering
    \includegraphics[width=.95\linewidth]{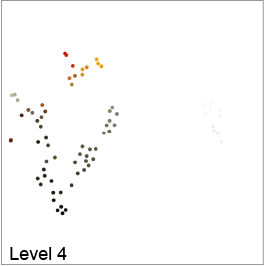}
    \label{sup:fig:bus_geo:emb:4}
  \end{subfigure}%
  \hfill
  \begin{subfigure}{0.14\textwidth}
    \centering
    \includegraphics[width=.95\linewidth]{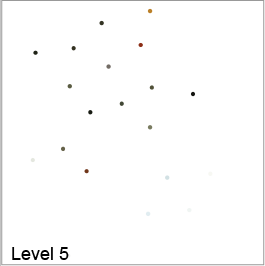}
    \label{sup:fig:bus_geo:emb:5}
  \end{subfigure}%
  \hfill
  \begin{subfigure}{0.14\textwidth}
    \centering
    \includegraphics[width=.95\linewidth]{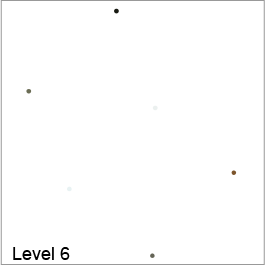}
    \label{sup:fig:bus_geo:emb:6}
  \end{subfigure}
 \caption{\textbf{Geodesic distance approach:} %
 RGB image of a bus (top left) and 6 levels of abstraction. Superpixels are recolored with the average color of all the image pixels they contain. Below, embeddings of each level using the same coloring. 
 The embeddings starting from level 1 are initialized based on the previous embedding.
 Numbers of components: 19090 (166x115 pixels), 5518, 1433, 356, 86, 21, 6. 
 Photograph (downscaled by factor 3) of the bus from Fig. 6 in \cite{Wei2018SuperpixelHierarchy}, available at \url{https://arxiv.org/abs/1605.06325}.%
 }
  \label{sup:fig:bus_geo}
\end{figure*}

\begin{figure*}
    \centering
    \includegraphics[width=\linewidth]{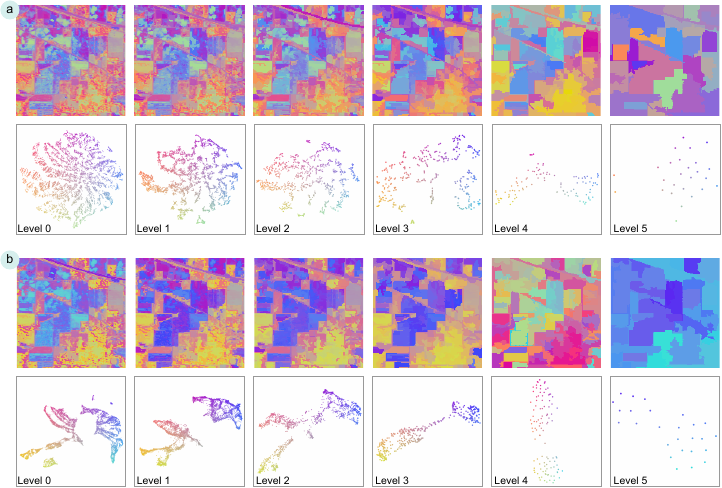}
    {\phantomsubcaption\label{fig:IndianPinesSmall:a}}%
    {\phantomsubcaption\label{fig:IndianPinesSmall:b}}%
    \vspace{-5mm}
    \caption{\textbf{Indian Pines:} %
    Embeddings and image space recolorings (using a 2D colormap as shown in \ref*{fig:IndianPinesLargeExploration:f}) for (a) t-SNE and (b) UMAP probabilities. 
    The embeddings starting from level 1 are initialized based on the previous embedding.
    }%
    \label{fig:IndianPinesSmall}%
\end{figure*}

\end{document}